\documentclass[accepted]{uai2026} 
                        
\usepackage[american]{babel}
\usepackage{graphicx}
\usepackage{color}
\usepackage{graphicx}
\usepackage{subcaption}
\usepackage{algorithm}
\usepackage[noend]{algorithmic}
\usepackage{minted}
\usepackage{verbatim}
\usepackage{amssymb}

\usepackage{natbib} 
\usepackage{mathtools} 
\usepackage{booktabs} 
\usepackage{tikz} 

\title{CoVAE: correlated multimodal generative modeling}

\author[1]{Federico Caretti}
\author[1]{\href{mailto:<gsanguin@sissa.it>?Subject=CoVAE}Guido Sanguinetti}

\affil[1]{%
    Scuola Internazionale Superiore di Studi Avanzati\\
    Trieste, Italy
}

\begin{document}
\maketitle

\begin{abstract}
Multimodal Variational Autoencoders have emerged as a popular tool to extract effective representations from rich multimodal data. However, such models rely on fusion strategies in latent space that destroy the joint statistical structure of the multimodal data, with profound implications for generation and uncertainty quantification. In this work, we introduce Correlated Variational Autoencoders (CoVAE), a new generative architecture that captures the correlations between modalities. We test CoVAE on a number of real and synthetic data sets demonstrating both accurate cross-modal reconstruction and effective quantification of the associated uncertainties.
\end{abstract}

\section{Introduction} \label{sec:introduction}
Learning compact representations of complex data is a central task in data science. Variational Autoencoders (VAEs, \cite{kingma2013auto}) have emerged as a popular choice in many applications. While overshadowed by more modern architectures in fields such as vision and language, they remain strongly popular in less data-rich areas, such as biomedical applications\citep{lopez2018deep} \citep{ashuach2023multivi} \citep{wei2020recent}, and scientific domains\citep{noh2019inverse} \citep{iten2020discovering} \citep{palma2026data} due to their ease of  interpretation and the natural uncertainty estimates that arise from their fitting procedure. Additionally, their generative nature, and the easy handling of conditioning on additional covariates, have contributed to make VAEs perhaps the most widely used unsupervised deep learning architecture in scientific applications.


Modern data sets often contain multiple, complementary observations for each object. The best studied example by far consists of images and related text, an example of such commercial importance that several methods have been specifically engineered for these two modalities. In science, multiple modalities are often commonplace; this has led in recent years to a flourishing of multi-modal VAE architectures (of which many have been recently reviewed and implemented in \cite{senellart2025multivae}) which adopt ingenious diverse strategies towards the common goal of integrating the different modalities in latent space.

While multimodal VAEs are a major success story, they face a fundamental conundrum: in many cases, it is highly desirable to perform inference even when only a subset of all modalities is available, therefore identifying a latent representation also in the case of missing data. This implies that separate modalities need to be independently encoded; integration is therefore performed in latent space, usually by fusing the different representations into a consensus one. This implies that multiple modalities are then decoded from a single latent point, and therefore the reconstructed modalities will necessarily be deterministically related, and have maximal mutual information. This subtle issue implies that the joint statistics of the generated multimodal data will necessarily not respect the original correlation structure, leading to profound implications for synthetic data generation and uncertainty estimation. This problem is particularly acute in scientific applications, where each modality may be affected by a multiplicity of sources of variation, and where correlations between different modalities may therefore vary widely.
 

\begin{figure}[h]
\centering

\begin{subfigure}[b]{0.49\linewidth}
    \centering
    \includegraphics[width=\linewidth]{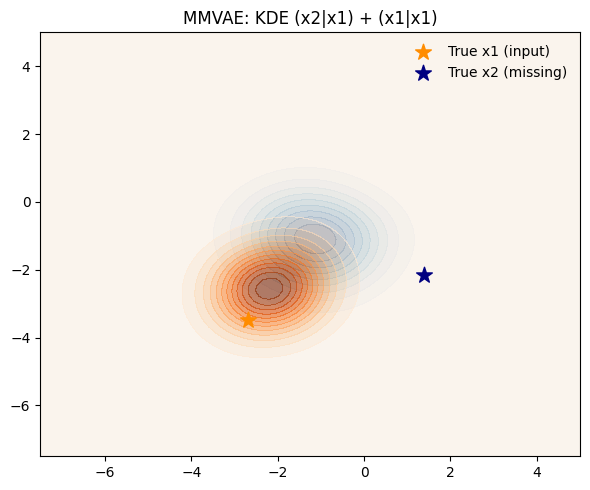}
    \label{subfig:kde_mmvae}
\end{subfigure}\hfill
\begin{subfigure}[b]{0.49\linewidth}
    \centering
    \includegraphics[width=\linewidth]{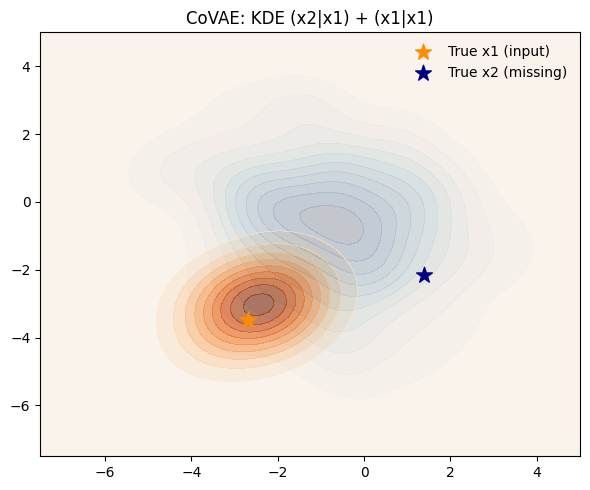}
    \label{subfig:kde_covae}
\end{subfigure}

\caption{When only one modality is available, common methods, such as Product-of-Experts\cite{wu2018multimodal} (left) erroneusly assign the same uncertainty to both modalities. On the contrary, CoVAE (right) correctly assigns a wider posterior to the missing modality. In this case, the two modalities have a linear correlation coefficient $\rho=0.5$}
\label{fig:linear_kde}
\end{figure}

To address this issue, we propose CoVAE, a multimodal Variational Autoencoder architecture that learns a multivariate, non-diagonal Gaussian structure within the latent space, therefore effectively storing the correlations between the modalities. The non-diagonal structure of the model also enables us to easily perform inference for unobserved modalities by sampling from the correct conditional distribution, therefore obtaining realistic estimates of uncertainty that account for partial correlations among modalities. We show that this simple approach performs remarkably effectively on a range of synthetic and real data sets, both in terms of reconstruction and generation of missing modalities.

\subsection{Related work}

Multimodal VAEs are commonly\citep{senellart2025multivae} divided into three categories: \textit{aggregated models} that extract information from all modalities and then combine them together, employing strategies such as Mixtures or Products of Experts; \textit{joint models} that employ a joint encoder that stacks all modalities together, complemented by unimodal encoders for the cases of missing modalities; \textit{coordinated models} that do not attempt to model the joint probability distribution, employing separate latent spaces instead and linking them through similarity constraints.

Among the aggregated models, the two most popular and intuitive strategies are Product-of-Experts (PoE), employed in MVAE \citep{wu2018multimodal} and MVTCAE \citep{hwang2021multi}, and Mixture-of-Experts (MoE), as in MMVAE \citep{shi2019variational}. The two different strategies can be interpreted as assigning power of veto to each expert or weighting their votes equally (although easy extensions of MMVAE allow for weighted means). Building upon them, \cite{sutter2021generalized} suggested that taking a Mixture-of-Product-of-Experts balances the benefits of both models while being an equally valid approximation of the posterior. 

The first of the joint VAE family has historically been JMVAE \citep{suzuki2016joint}, in which a joint encoder is trained to approximate the joint posterior distribution, and single-modality encoders are trained to approximate the same distribution by minimizing a Kullback-Leibler divergence among the individual representations. More recent works in this family include TELBO\citep{vedantam2018generative}, which modifies the loss function and training regime while employing essentially the same architecture as JMVAE, and JNF-DCCA \citep{senellart2023improving}, which enriches the unimodal posteriors with normalizing flows.

These approaches suffer from a collapse in the joint statistics: because the different modalities are eventually summarised in a single latent distribution, the decoded modalities will be deterministically related. As a consequence, generated data will have maximal mutual information between different modalities, clearly unlikely to be the case in real data. The practical consequences of this are illustrated in Fig.\ref{fig:linear_kde}, which illustrates a case where reconstruction is performed based on a single modality while the second is imputed: here we see that an aggregation strategy based on PoE will provide a significant underestimate of the variance associated with an unseen modality (left panel), a problem that is corrected by our approach CoVAE (right panel).

More recently, it has been proposed to use different latent spaces for each modality, paired with a shared latent space that aims at capturing the common information. Both MMVAEPlus\citep{palumbo2023mmvae+} and DMVAE\citep{lee2021private} project the data in both a shared latent space (through MoE and PoE respectively) and modality specific latent spaces and then decode this augmented latent space. In principle, this approach allows for the modeling of correlations; however, we show in Sec.\ref{sec:experiments} that in practice the current implementations of these models do not achieve the goal of preserving inter-modality statistical dependencies.

\section{Methods} \label{sec:methods}

\subsection{Variational Autoencoders and their multimodal extensions} \label{sec:vaes}
Variational Autoencoders (VAEs) are a class of latent-variable generative models. Initially introduced by \cite{kingma2013auto}, they still see widespread use because of their versatility in modeling distributions $p(\mathbf{x})$ over complex, high dimensional data $\mathbf{x}\in\mathbb{R}^D$ as nonlinear transformations (parameterised by a neural network with parameters $\theta$) of a simple distribution over  latent variables $\mathbf{z}\in\mathbb{R}^d\quad d<<D$, usually a simple standard normal. Since the evaluation of the marginal likelihood of the observed data points is intractable when the mapping between latent and observed variables is nonlinear, VAEs instead  maximize the Evidence Lower Bound (ELBO).

\begin{equation}
\label{eq:elbo}
\mathcal{L}(\theta,\phi;\mathbf{x})
  = \mathbb{E}_{q_\phi(\mathbf{z}\mid\mathbf{x})}\!\bigl[\log p_\theta(\mathbf{x}\mid\mathbf{z})\bigr]
    - \operatorname{KL}\!\bigl[q_\phi(\mathbf{z}\mid\mathbf{x}) \,\|\, p(\mathbf{z})\bigr].
\end{equation}
Here $q_\phi(\mathbf{z}\mid\mathbf{x})$ is an \emph{encoder} (inference network), 
$p_\theta(\mathbf{x}\mid\mathbf{z})$ is a \emph{decoder} (generative network) and 
$p(\mathbf{z})$ is typically a standard Gaussian prior. The reparameterization trick allows the backpropagation of the gradients even through the sampling step, by parameterizing samples from $q_{\phi}$ as $\mathbf{z} = \boldsymbol{\mu}_\phi + \boldsymbol{\sigma}_\phi \odot \boldsymbol{\epsilon}$, with $\boldsymbol{\epsilon} \sim \mathcal{N}\left(\mathbf{0}, \mathbf{I}\right)$.


In the presence of multiple modalities, VAEs can be trivially extended by simply concatenating modalities. However, this naive strategy has the major drawback that data points with a missing modality cannot be utilized; instead, recent research has focused on architectures which have separate encoders/ decoders for different modalities, and then devise integration strategies in a shared latent space, e.g. by treating the different modalities as different experts and taking either the sum or the product of their contributions to the posterior.
$$
q_\phi(\mathbf{z} \mid \mathbf{x}_S)
\;\propto\;
\;
\left\{
\begin{array}{ll}
p(\mathbf{z}) \displaystyle\prod_{m \in S} q_{\phi_m}(\mathbf{z} \mid \mathbf{x}_m) & \quad\text{(PoE)}\\[12pt]
\dfrac{1}{|S|} \displaystyle\sum_{m \in S} q_{\phi_m}(\mathbf{z} \mid \mathbf{x}_m) & \quad\text{(MoE)}
\end{array}
\right.
$$

This strategy inevitably leads to generated data which are deterministically related, and to potentially serious errors in quantifying the total uncertainty in the case of missing modalities, as discussed before and illustrated in Figure \ref{fig:linear_kde}.




\subsection{Correlated VAEs (CoVAE)}

As discussed before, standard multimodal VAEs excel at approximating complex multimodal distributions in terms of compact representations. However, to do so, they collapse the representations of the multiple modalities in a single latent point, hence introducing spurious deterministic dependencies between modalities in generated data. Therefore, when used to generate missing modalities, they will usually be overconfident (e.g. providing sharp images), regardless of the actual information the remaining modalities may contain with respect to the missing one. The core idea of Correlated VAEs (CoVAE) is to approach this problem by encoding correlations between modalities in a non-diagonal prior, and learning a joint encoder able to produce data points with the correct amount of correlation through the latent space regularization.

\begin{figure}[h]
\centering
\includegraphics[width=0.8\linewidth]{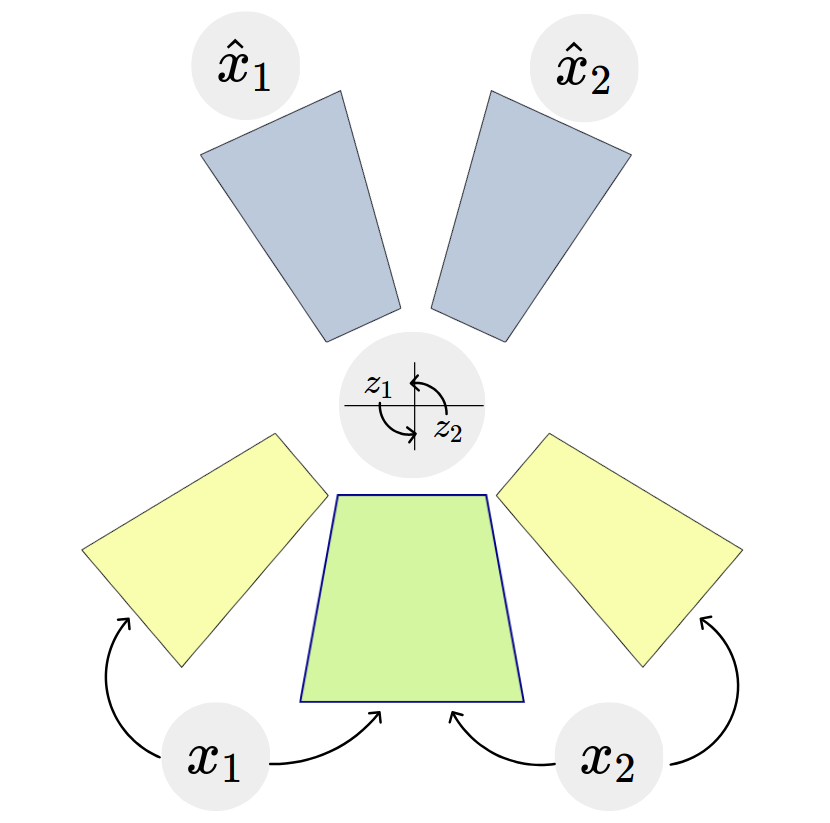}
\caption{Graphical representation of CoVAE: in the case of two modalities, during training }
\end{figure}
\paragraph{CoVAE architecture}
Let us assume that the data is observed through $K$ distinct modalities. CoVAE postulates that each modality is encoded separately into a $d$-dimensional latent space through an encoder network $q_{\phi_k}(\mathbf{z}_k \mid \mathbf{x}_k)$, defining the distribution for the latent variable associated to the $k$-th modality\footnote{In principle, each modality could be encoded into spaces of different dimension, we omit this option in the discussion to avoid overburdening the notation.}. This encoder network is a standard VAE encoder with diagonal covariance.

We also define the latent variable $\mathbf{z}\in\mathbb{R}^{dK}=\times_{k=1}^K\mathbb{R}^d$ obtained by concatenating all the modality-specific latent variables. This concatenated variable is given a multivariate normal prior $p(\mathbf{z}) = \mathcal{N}(\mathbf{0}, \Sigma_{\mathrm{prior}})$. The encoder network for the concatenated latent variable gives rise to a full covariance multivariate Gaussian  

$$q_\phi(\mathbf{z}\mid\mathbf{x})=\mathcal{N}(\boldsymbol{\mu},\, \Sigma_{\mathrm{joint}}).$$
We choose a Cholesky decomposition as the default parameterization in order to guarantee the symmetry and positive definiteness of the covariance matrix: \[\Sigma_{\mathrm{joint}} \;=\; \mathbf{L} \mathbf{L}^\top \;+\; \varepsilon \mathbf{I}\] 
where $\mathbf{L}$ is a lower triangular matrix whose entries are parameterized by a set of weights $\mathbf{w}$.
We observe that the model belongs to the family of joint models: when all modalities are present, a joint encoder parameterizes $q_\phi(\mathbf{z}\mid\mathbf{x})$ through $\boldsymbol{\mu}$ and $\mathbf{w}$.

In the absence of data for a subset of modalities, the inference of the missing latent variables can be performed by using the modality specific encoders and  conditioning on the projected values under the prior $p(\mathbf{z}) = \mathcal{N}(\mathbf{0}, \Sigma_{\mathrm{prior}})$. Denoting by $\mathcal{O}$ the set of observed modalities and by $\mathcal{M}$ the set of missing ones, the conditional distribution is:

\begin{equation}\begin{split} \label{eq:conditioning}
\mathbf{z}_{\mathcal{M}} \mid &\mathbf{z}_{\mathcal{O}} \sim\\
&\mathcal{N}\Big(
\Sigma_{\mathcal{M}\mathcal{O}}\,\Sigma_{\mathcal{O}\mathcal{O}}^{-1}\,\mathbf{z}_{\mathcal{O}},\;
\Sigma_{\mathcal{M}\mathcal{M}}-\Sigma_{\mathcal{M}\mathcal{O}}\,\Sigma_{\mathcal{O}\mathcal{O}}^{-1}\,\Sigma_{\mathcal{O}\mathcal{M}}
\Big)
\end{split}\end{equation}

where subscripts on $\Sigma$ refer to the corresponding blocks of $\Sigma_{\mathrm{prior}}$.

\paragraph{Training CoVAEs}

We train the joint part of CoVAE to minimize the following loss function:
\begin{multline}\label{eq:joint_loss}
\mathcal{L}_{\mathrm{joint}} = -\sum_{\mathbf{x}\in\mathcal{D}} \bigg\{
\mathbb{E}_{q_\phi(\mathbf{z}\mid\mathbf{x})}\!\bigg[\sum_{k=1}^{K}\log p_{\theta_k}(\mathbf{x}_k \mid \mathbf{z}_k)\bigg] \\
- \beta\, \mathrm{KL}\!\Big(q_\phi(\mathbf{z}\mid\mathbf{x}) \;\Big\|\; \mathcal{N}\!\big(\mathbf{0},\, \Sigma_{\mathrm{prior}}\big)\Big)\bigg\},
\end{multline}

as in standard VAEs.

As in standard practice for multimodal VAEs, we also train the unimodal encoders: for each modality $k$, the unimodal encoder produces $\mathbf{z}_k \sim q_{\phi_k}(\mathbf{z}_k \mid \mathbf{x}_k)$. The remaining latents are sampled from the prior conditional $\mathbf{z}_{-k} \sim p(\mathbf{z}_{-k}\mid\mathbf{z}_k)$ using Eq.~\eqref{eq:conditioning} with $\mathcal{O}=\{k\}$ and $\mathcal{M}=\{-k\}$, and all modalities are reconstructed:
\begin{multline}\label{eq:cond_loss}
\mathcal{L}_{\mathrm{cond},k} = -\sum_{\mathbf{x}\in\mathcal{D}} \bigg\{
\mathbb{E}_{\substack{q_{\phi_k}(\mathbf{z}_k\mid\mathbf{x}_k) \\ p(\mathbf{z}_{-k}\mid\mathbf{z}_k)}}\!\bigg[\sum_{j=1}^{K}\log p_{\theta_j}(\mathbf{x}_j \mid \mathbf{z}_j)\bigg] \\
- \beta\,\mathrm{KL}\!\Big(q_{\phi_k}(\mathbf{z}_k\mid\mathbf{x}_k) \;\Big\|\; \mathcal{N}\!\big(\mathbf{0},\, \Sigma_{\mathrm{prior}}^{(kk)}\big)\Big)\bigg\}.
\end{multline}

Therefore, for every training step over a minibatch, we both train the model's joint encoder by minimizing objective~\eqref{eq:joint_loss} and the single modality encoders by sampling from Eq.~\eqref{eq:conditioning} and minimizing Eq.~\eqref{eq:cond_loss}.  We summarize in pseudocode the training process for CoVAE with $K$ modalities and one joint encoder in Table \ref{alg:pseudocode}. 

\begin{algorithm}[t]
\caption{Pseudocode of CoVAE training}
\label{alg:pseudocode}
\begin{algorithmic}[1]
\STATE \textbf{Pre-training:} run Deep CCA on $\{q_{\phi_k}\}_{k=1}^K$ to obtain $\Sigma_{\mathrm{prior}}$; freeze $\Sigma_{\mathrm{prior}}$
\medskip
\FOR{each minibatch $\mathcal{B}$}
\STATE $\boldsymbol{\mu}_\mathbf{x}, \mathbf{L}_\mathbf{x} \leftarrow q_\phi(\mathbf{x}) \;\;\forall\, \mathbf{x}\in\mathcal{B}$; \quad $\mathbf{z} \sim \mathcal{N}(\boldsymbol{\mu}_\mathbf{x},\, \mathbf{L}_\mathbf{x} \mathbf{L}_\mathbf{x}^\top\!+\varepsilon \mathbf{I})$ 
\STATE $\mathcal{L}_{\mathrm{joint}} \leftarrow -\frac{1}{|\mathcal{B}|}\sum_\mathbf{x}\big\{ \sum_k \log p_{\theta_k}(\mathbf{x}_k\mid\mathbf{z}_k) - \beta\,\mathrm{KL}\big(q_\phi(\mathbf{z}\mid\mathbf{x})\,\|\,\mathcal{N}(\mathbf{0},\Sigma_{\mathrm{prior}})\big)\big\}$
\FOR{$k=1,\dots,K$}
\STATE $\mathbf{z}_k \sim q_{\phi_k}(\mathbf{z}_k\mid\mathbf{x}_k)$;\quad $\mathbf{z}_{-k} \sim p(\mathbf{z}_{-k}\mid\mathbf{z}_k)$ \COMMENT{Eq.~\eqref{eq:conditioning} with $\mathcal{O}=\{k\}$}
\ENDFOR
\STATE $\mathcal{L}_{\mathrm{cond}} \leftarrow -\frac{1}{|\mathcal{B}|}\sum_k\sum_\mathbf{x}\big\{ \sum_j \log p_{\theta_j}(\mathbf{x}_j\mid\mathbf{z}_j) - \beta\,\mathrm{KL}\big(q_{\phi_k}(\mathbf{z}_k\mid\mathbf{x}_k)\,\|\,\mathcal{N}(\mathbf{0},\Sigma_{\mathrm{prior}}^{(kk)})\big)\big\}$
\STATE Update $\theta,\phi,\{\phi_k\}$ with $\nabla\!\big(\mathcal{L}_{\mathrm{joint}} + \lambda\,\mathcal{L}_{\mathrm{cond}}\big)$
\ENDFOR
\end{algorithmic}
\end{algorithm}

The weights of the prior's covariance matrix, which store the cross-modality correlations, can in principle be optimized during training. However, in practice we found more effective to first learn the correlations through Deep-CCA\cite{andrew2013deep}, simultaneously pre-training the single modality encoders, and freezing the prior's weights. A brief comparison of the two options is presented in Appendix\ref{supp:exp}.

As is the case for other joint models, CoVAE in principle requires to train every subset of combinations of $k$ modalities to perform inference in the case of absence of one or more modalities, requiring $2^K$ encoders to be trained.  In practice, the number of modalities encountered in many practical applications is small; we acknowledge though that this may be a limitation in some cases.

\paragraph{Generating from CoVAE}

Unconditional generation in VAEs is in principle possible by decoding directly latent values sampled from the prior. In practice, this often leads to nonsensical results, as the posteriors can be concentrated far from the prior. For these reasons, standard practice involves sampling from an approximation of the aggregated posterior obtained by mixing all the training posterios, for example utilising Gaussian Mixture Models and Normalizing Flows\citep{rezende2015variational}.

Conditional generation in multimodal VAEs is usually done by encoding the observed modalities and then decoding samples from the relative posteriors. This procedure is what collapses the joint statistics of the generated data. In CoVAE, instead, the observed modalities are encoded and the single modality samples are then used to generate from the conditional prior a sample point to be decoded for the missing modality using eq. \eqref{eq:conditioning}.
\section{Experiments} 

\label{sec:experiments}

We test our model against competitive alternatives, namely JMVAE \citep{suzuki2016joint}, MVAE \citep{wu2018multimodal}, MMVAE  \citep{shi2019variational}, MVTCAE \citep{hwang2021multi} , MoPoE \citep{sutter2021generalized}, MMVAE+\citep{palumbo2023mmvae+} and DMVAE \citep{lee2021private}. All the models are called through their implementations in the package MultiVAE \citep{senellart2025multivae}, and in each experiment we standardize the architecture across the models in order to attribute the difference in performance to the multimodal VAE architecture used. The NLLs, both joint and conditional, are calculated using the IWAE estimator \citep{burda2015importance}. For DMVAE, the tables do not report the negative log-likelihoods because it was orders of magnitude higher than the other models, likely for an error in the implementation of the library.
\subsection{Synthetic datasets}\label{sec:synthetic_data}
\begin{figure}[h]
\centering
\includegraphics[width=\linewidth]{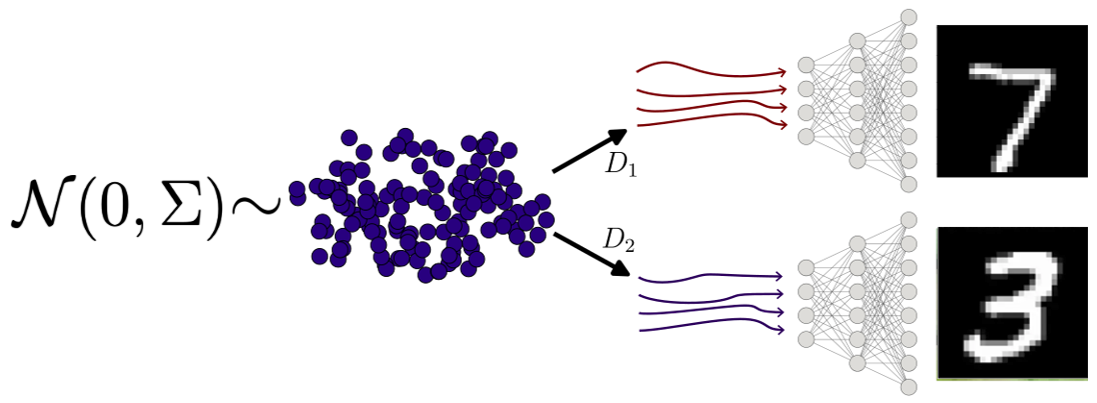}
\caption{Schematic process of the data generation, assuming classes coming from MNIST}\label{fig:synthGen}
\end{figure}

We first test CoVAE on a simulated bimodal data set of pairs of MNIST digits with controlled correlation.
To generate the synthetic datasets we use the following procedure: first, we draw samples of concatenated latent variables from a multivariate Gaussian distribution with the desired correlation. Next, we use pre-trained decoders to generate realistic samples of digits. The schematic procedure is illustrated in Fig. \ref{fig:synthGen}.

Given $D_1$ and $D_2$ the dimensions of the latent spaces of the two modalities, and $0 \leq \rho < 1$ the desired correlation, we draw samples from:

\begin{equation} \label{eq:covariance_generation}
\Sigma \;=\;
\begin{pmatrix}
\mathbf{I}_{D_1}               & \rho\,\mathbf{I}_{D_1,D_2} \\[1ex]
\rho\,\mathbf{I}_{D_2,D_1}   & \mathbf{I}_{D_2}
\end{pmatrix}
\in\mathbb{R}^{(D_1+D_2)\times(D_1+D_2)}
\end{equation}
where $\mathbf{I}_{D_k}$ denotes the $D_k \times D_k$ identity matrix and
$\mathbf{I}_{D_1,D_2} \in \mathbb{R}^{D_1 \times D_2}$ is the rectangular matrix
with ones on the pseudo-diagonal and zeros elsewhere, i.e.\
$(\mathbf{I}_{D_1,D_2})_{ij} = \delta_{ij}$ for $i=1,\dots,D_1$, $j=1,\dots,D_2$.

The matrix $\Sigma$ is guaranteed to be positive semidefinite by construction and, therefore, a valid covariance matrix. Furthermore, it is guaranteed that for each modality, the marginalization over the other modalities results in the standard normal distribution:

$$z_{k} \sim \mathcal{N}(0,I_{D_{k}})$$

After sampling $\mathbf{z} \sim \mathcal{N}(\mathbf{0}, \Sigma)$, we generate data
in the observed space using pre-trained decoders. For each modality $k$, we first
train a VAE to obtain a decoder $p_{\theta_k}(\mathbf{x}_k \mid \mathbf{z}_k)$.
Optionally, we compose the sampling with a
masked autoregressive flow~\citep{papamakarios2017masked} $f_{\psi_k}$, trained to transform the standard
normal marginal $\mathcal{N}(\mathbf{0}, \mathbf{I}_{D_k})$ into the aggregate
posterior of the pre-trained VAE, yielding
\begin{equation}\label{eq:synthetic_generation}
\mathbf{x}_k \sim p_{\theta_k}\!\bigl(\mathbf{x}_k \mid f_{\psi_k}(\mathbf{z}_k)\bigr).
\end{equation}

We do not artificially link classes across modalities (i.e., we do not constrain the two digits in the pair to be related as digits). In principle, this could trivially be possible by employing a conditional decoder and conditioning on the desired class.

\begin{figure}[h]
\centering
\includegraphics[width=0.7\linewidth]{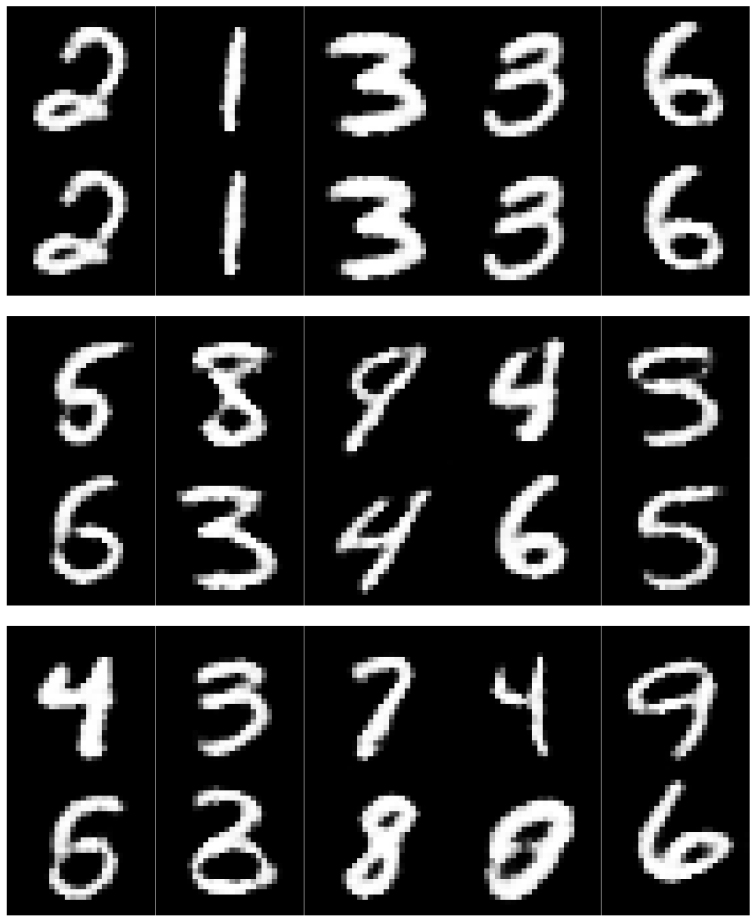}
\caption{Examples of synthetic datasets in which both modalities are from the MNIST dataset and latent dimensions $D_1=D_2=10$. From top to bottom, $\rho=0.99, 0.7, 0.05$}\label{fig:examples_MNIST}
\end{figure}

We create datasets of 20000 pairs of images, generated from a $D_1=D_2=10$ dimensional latent space with different levels of correlation following the described procedure. Examples of data pairs at different correlation levels are shown in Fig. \ref{fig:examples_MNIST}.

\begin{figure}[h] 
    \centering
    \includegraphics[width=\linewidth]{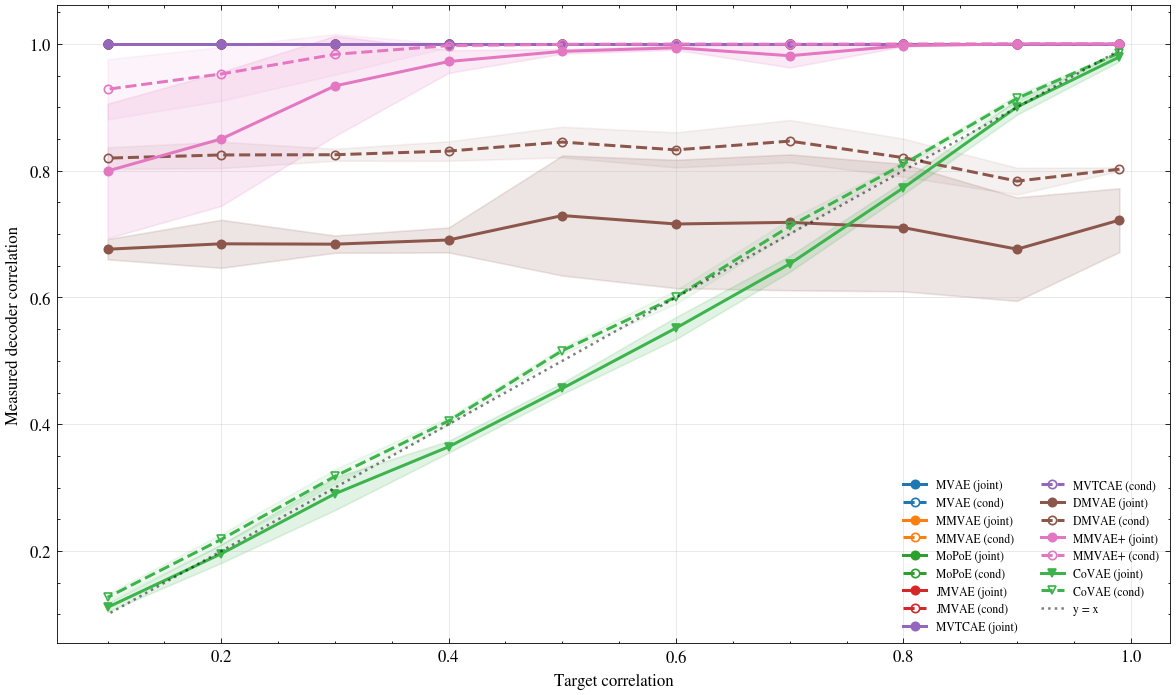}
    \caption{Linear correlations measured at the level of the input to the decoders, both in the case of joint and conditional reconstructions. The intervals around the measured values represent $\pm1\sigma$}
    \label{fig:correlations}
\end{figure}

\paragraph{Results}

\begin{figure*}[h]
    \centering
        \includegraphics[width=0.95\linewidth]{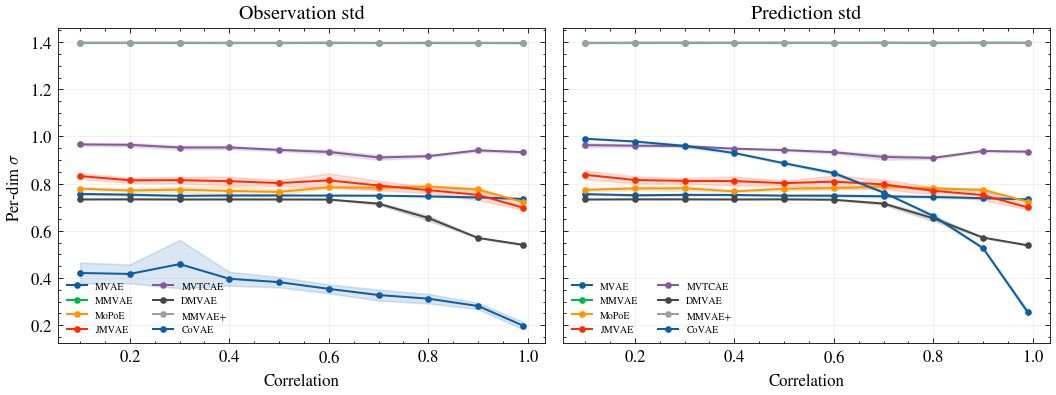}
    \caption{Average standard deviation per dimension for the reported models in a conditional generation setup. On the left, the standard deviation of the distribution of the conditioning modality; on the right, the standard deviation of the distribution of the missing modality}
    \label{fig:stds}
\end{figure*}
We benchmarked CoVAE against all competitors using the correct latent dimension $D=20$ for all models. All results report averages and standard deviations across 5 replications.

As a first test, we check whether the models can reconstruct data with the correct statistical structure, i.e. with the right correlation, by measuring the linear correlation between the modalities in the first decoder layer. As expected, we observe that the CoVAE architecture is the only one able to reconstruct data with the correct amount of correlation: Fig. \ref{fig:correlations} shows that both the joint and the conditional generation reconstruct the correct level of correlation. All the other models fail at this task: this is obvious in the case of the single latent space models, which can only generate data with maximal mutual information, reflected in the correlation of 1 that we observe. More interestingly, DMVAE and MMVAEPlus have the ability to generate non-trivial correlations. However, because learning cross-modality correlation is not part of the training procedure of the models, the observed correlations in Fig. \ref{fig:correlations} is constant across all levels of true correlation, and depends more on the hyperparameters of the model than the actual correlation in the data.

We then considered the impact of correlations on uncertainty estimation in Fig. \ref{fig:stds}. Here the difference between CoVAE and all the other models is again stark: the uncertainty estimated by all other models does not depend on the level of correlation. Perhaps more surprisingly, it does not depend even on which modality has been observed: comparing the left and right panels Fig.\ref{fig:stds} we see that all models bar CoVAE assign the same uncertainty to reconstructed data regardless of whether the specific modality was observed or not!  CoVAE, instead, can recognize which modality has been observed and therefore assign a lower uncertainty to it, when compared to the unobserved one. Additionally, because CoVAE leverages correlations, uncertainties decrease the more we increase the correlation between modalities. This is true even in the observed modality, because training on correlated data leads to more effective information sharing, but it is even more prominent in reconstructing missing modalities, where correlation is rightly reflected in a strongly decreasing trend in the predictive standard deviation. We also highlight that, according to CoVAE, the uncertainty on the unobserved modality correctly starts, for $\rho \rightarrow 0$, at $1$, that is, the standard deviation of the prior. 

\begin{figure}[h]
    \centering
    \begin{subfigure}{\columnwidth}
        \centering
        \includegraphics[width=0.95\columnwidth]{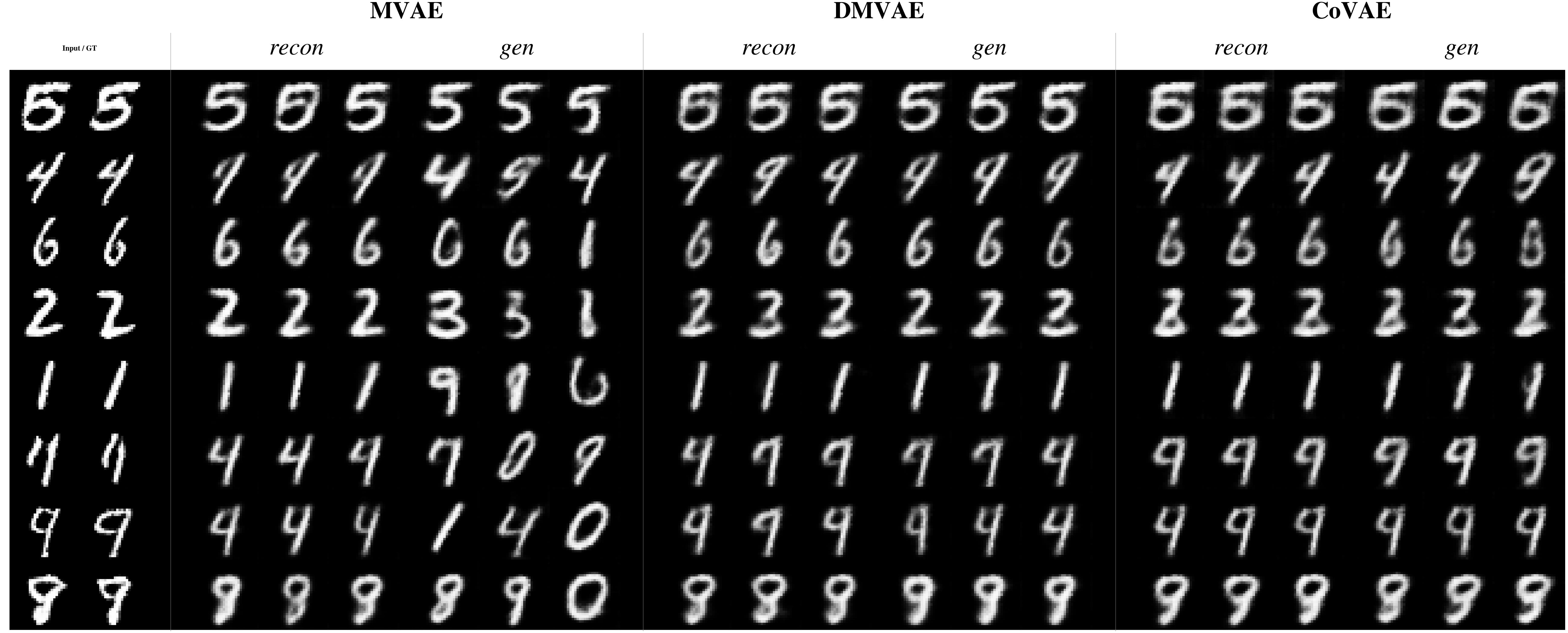}
    \end{subfigure}
    \vspace{0.5em}
    \begin{subfigure}{\columnwidth}
        \centering
        \includegraphics[width=0.95\columnwidth]{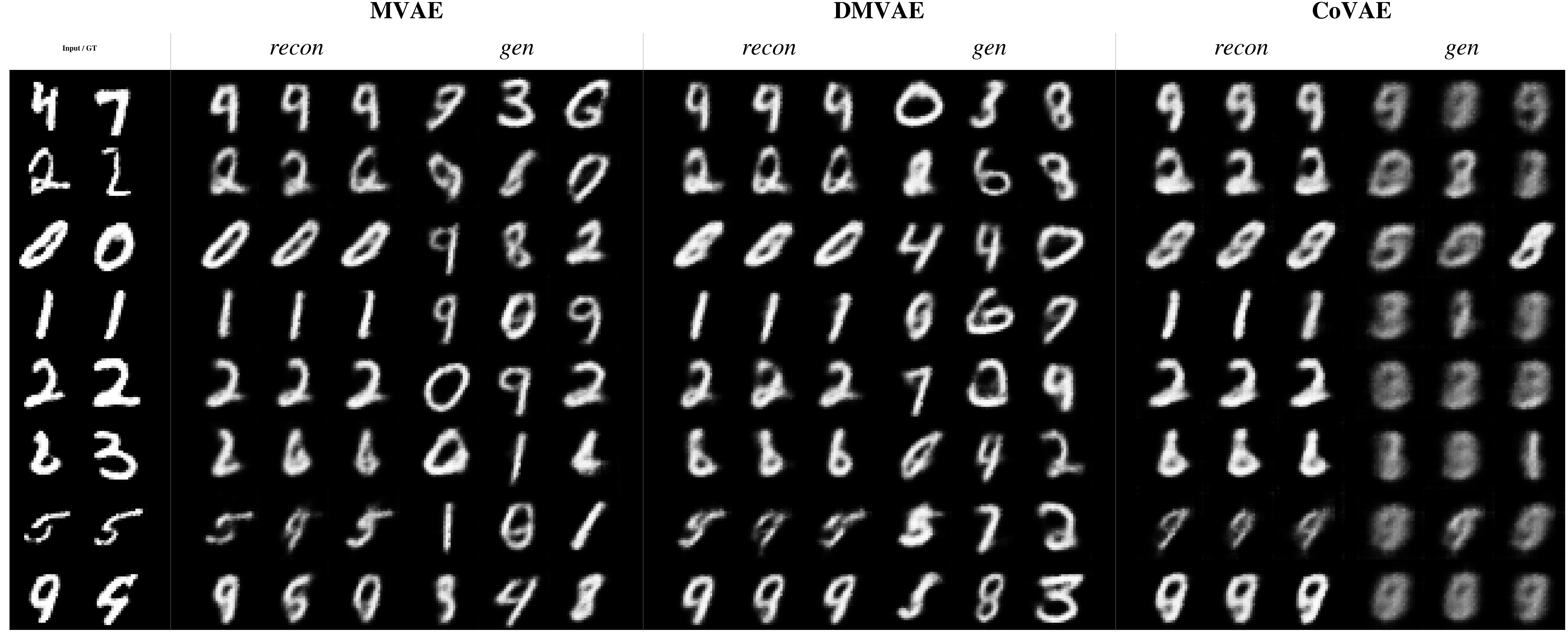}
    \end{subfigure}
    \caption{Conditional reconstruction of input modalities for MVAE (first row), DMVAE (second) and CoVAE (third) at $\rho=0.9$ (top) and $\rho=0.6$ (bottom). For each model, the first three images represent the generated input modality, followed by samples of the missing modality}
    \label{fig:example_conditional_generations}
\end{figure}

We provide some examples of the impact of correlation in the space of images in Fig.\ref{fig:example_conditional_generations}. Here we compare the reconstructions and conditional generations in a few examples of three models, MVAE, DMVAE and CoVAE. The top panel shows examples with high correlations; here both DMVAE and CoVAE perform very well with  sharp images of the correct digit. MVAE on the contrary exhibits more variability in conditional generation, often generating wrong digits. The difference is more marked at intermediate correlations: here MVAE and DMVAE generate sharp digits which are generally wrong. CoVAE generates fainter digits but more recognisably right: the correlations help CoVAE locate the conditional samples closer to the correct posterior, albeit the resulting uncertainty means that they may lie slightly outside the manifold of digits.

\begin{figure}[t]
    \centering
    \begin{minipage}[t]{0.48\linewidth}
        \centering
        \includegraphics[width=\linewidth]{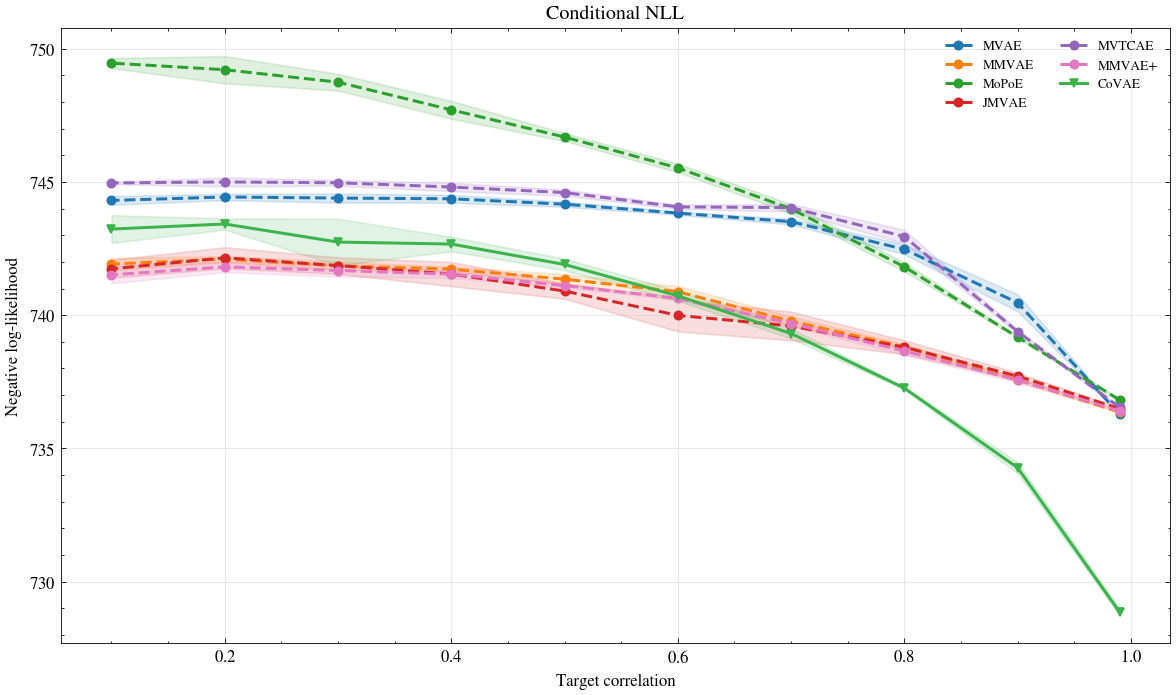}
    \end{minipage}
    \hfill
    \begin{minipage}[t]{0.48\linewidth}
        \centering
        \includegraphics[width=\linewidth]{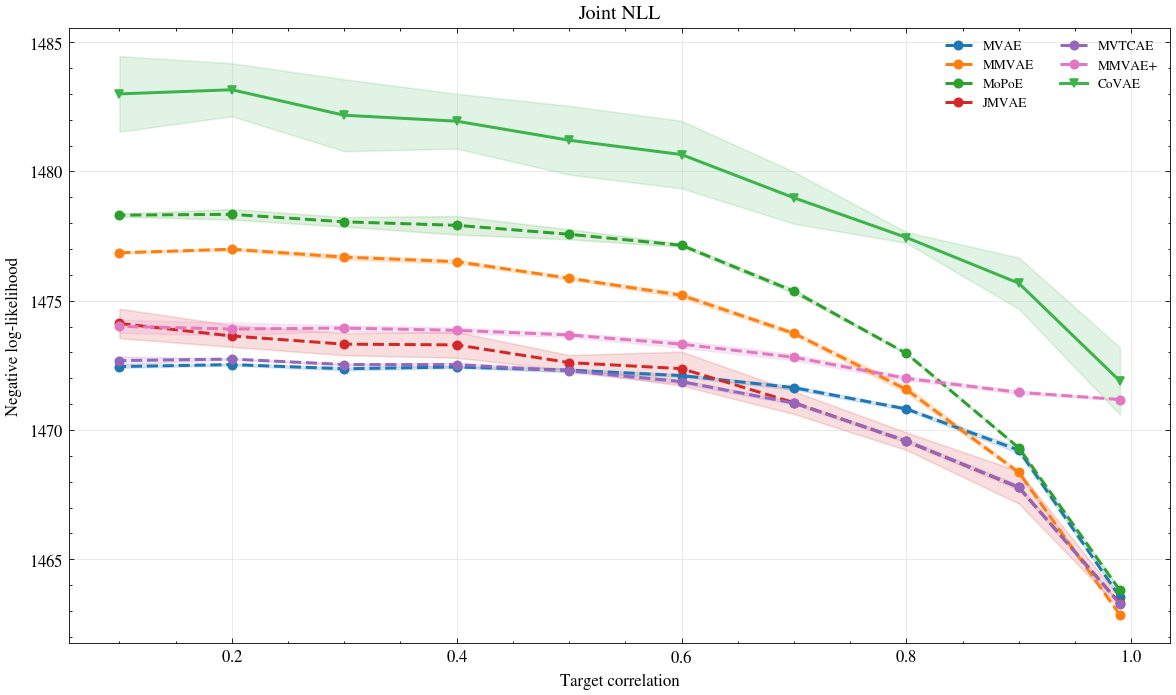}
    \end{minipage}
    \caption{Negative log-likelihoods for different models at varying levels of correlation.}
    \label{fig:nll_comparison}
\end{figure}
Finally, we compare the various models in terms of (test) joint and conditional negative log-likelihood (NLL) in Fig.\ref{fig:nll_comparison}. Here, CoVAE pays an entropic price for having correlated Gaussians in latent space (we decompose and show the contributions to the NLL in Appendix\ref{supp:jointNLL}). Nevertheless, we see a marked decrease in both NLL and particularly joint NLL as we increase the correlation of the generated images, further evidence that CoVAE is able to effectively leverage correlations across modality to improve learning and generation.

\subsection{Biomedical dataset}

\paragraph{Setup description} We test all the aforementioned models on a multimodal biological dataset imported from \cite{yang2025mlomics}, originally created within the TCGA project. We chose the Pan-Cancer dataset containing 8314 samples and 2 paired modalities: 3217 mRNA features and 383 miRNA features. All samples are paired and are labeled with one of 32 cancer types. No feature selection is applied, and all the features are scaled to have 0 mean and unit variance. The Neural Network architectures are the same across all the models, with both encoders and decoders being fully connected layers and $D=dim(\mathbf{z})=32$. For CoVAE, we set $D=32$ and split equally the latent variables among the two modalities. For models requiring private latent spaces, we set $D_{private}=D/2$ and then run the classifiers on the shared latent space, after testing that this increases the performance of all metrics. Details on the implemented metrics are in Appendix\ref{supp:exp}.

\paragraph{Results}

The prior correlation learned by CoVAE  yields $\rho = 0.78$, indicating a strong linear dependence between the latent representations of mRNA and miRNA. Although there is no ground truth available for this dataset, this high value informs us that we can expect good results on the cross-modality tasks.

Table\ref{tab:joint_results} summarizes the performance of the different models on some popular metrics that evaluate the quality of the cancer type classification. The cancer type classifiers operate in the latent space; therefore, these metrics value inference more than reconstruction quality. Interestingly, although MoPoE performs better on all the latent space classification metrics, its NLL is the worst among the tested models. The performance of the CoVAE model is slightly worse than that of the models that employ a Product-of-Experts strategy, and comparable to or slightly better than the other models. In any case, the cancer type classification results appear to be consistently good across all methods when all modalities are present (and in fact they are also similar when only one modality is available).

\begin{table*}[h]
\centering
\resizebox{0.8\textwidth}{!}{%
\begin{tabular}{lcccc}
\toprule
Model & Precision $\uparrow$ & NMI $\uparrow$ & ARI $\uparrow$ & Joint LL $\downarrow$ \\
\midrule
MVAE    & 0.901 ± 0.011$^{(3)}$ & 0.899 ± 0.011$^{(3)}$ & 0.844 ± 0.018$^{(3)}$ & 3839.297 ± 2.571$^{(3)}$ \\
MVTCAE  & 0.905 ± 0.010$^{(2)}$ & 0.903 ± 0.011$^{(2)}$ & 0.850 ± 0.017$^{(2)}$ & 3837.561 ± 2.893$^{(2)}$ \\
MMVAE   & 0.865 ± 0.012$^{(7)}$ & 0.857 ± 0.013$^{(7)}$ & 0.780 ± 0.020$^{(7)}$ & 3859.866 ± 4.449$^{(6)}$ \\
MoPoE   & \textbf{0.918 ± 0.009}$^{(1)}$ & \textbf{0.918 ± 0.009}$^{(1)}$ & \textbf{0.875 ± 0.015}$^{(1)}$ & 3865.613 ± 3.137$^{(7)}$ \\
JMVAE   & 0.892 ± 0.010$^{(5)}$ & 0.891 ± 0.010$^{(5)}$ & 0.831 ± 0.016$^{(5)}$ & 3846.290 ± 2.001$^{(4)}$ \\
DMVAE   & 0.832 ± 0.019$^{(8)}$ & 0.838 ± 0.018$^{(8)}$ & 0.738 ± 0.029$^{(8)}$ & — \\
MMVAE+  & 0.881 ± 0.013$^{(6)}$ & 0.880 ± 0.014$^{(6)}$ & 0.814 ± 0.024$^{(6)}$ & \textbf{3431.332 ± 2.735}$^{(1)}$ \\
CoVAE   & 0.893 ± 0.033$^{(4)}$ & 0.891 ± 0.036$^{(4)}$ & 0.834 ± 0.056$^{(4)}$ & 3856.761 ± 2.596$^{(5)}$ \\
\bottomrule
\end{tabular}
}
\caption{Comparison of multimodal VAE models on the pan-cancer mRNA/miRNA dataset on joint tasks.}
\label{tab:joint_results}
\end{table*}

\begin{table*}[h]
\centering
\resizebox{\textwidth}{!}{%
\begin{tabular}{lcccccc}
\toprule
Model & mRNA→miRNA MAE $\downarrow$ & miRNA→mRNA MAE $\downarrow$ & mRNA→miRNA Cls $\uparrow$ & miRNA→mRNA Cls $\uparrow$ & LL $\mid$ mRNA $\downarrow$ & LL $\mid$ miRNA $\downarrow$ \\
\midrule
MVAE    & 0.639 ± 0.033$^{(8)}$ & 0.563 ± 0.047$^{(5)}$ & 0.832 ± 0.029$^{(6)}$ & 0.784 ± 0.068$^{(7)}$ & 448.392 ± 1.849$^{(7)}$ & 3466.259 ± 10.359$^{(7)}$ \\
MVTCAE  & 0.623 ± 0.011$^{(7)}$ & 0.582 ± 0.019$^{(6)}$ & 0.690 ± 0.026$^{(8)}$ & 0.739 ± 0.027$^{(8)}$ & 444.148 ± 0.550$^{(4)}$ & 3462.294 ± 2.720$^{(6)}$ \\
MMVAE   & 0.465 ± 0.003$^{(2)}$ & 0.618 ± 0.081$^{(7)}$ & 0.786 ± 0.056$^{(7)}$ & 0.815 ± 0.045$^{(6)}$ & 442.770 ± 1.103$^{(3)}$ & 3445.171 ± 2.906$^{(3)}$ \\
MoPoE   & \textbf{0.460 ± 0.002}$^{(1)}$ & 0.406 ± 0.002$^{(1)}$ & 0.926 ± 0.008$^{(2)}$ & 0.927 ± 0.007$^{(2)}$ & 448.273 ± 0.583$^{(6)}$ & 3456.986 ± 3.752$^{(5)}$ \\
JMVAE   & 0.479 ± 0.001$^{(5)}$ & 0.441 ± 0.001$^{(3)}$ & \textbf{0.935 ± 0.007}$^{(1)}$ & \textbf{0.938 ± 0.007}$^{(1)}$ & 433.423 ± 0.284$^{(2)}$ & \textbf{3433.295 ± 2.204}$^{(1)}$ \\
DMVAE   & 0.575 ± 0.014$^{(6)}$ & 0.751 ± 0.017$^{(8)}$ & 0.891 ± 0.011$^{(5)}$ & 0.904 ± 0.010$^{(4)}$ & — & — \\
MMVAE+  & 0.468 ± 0.003$^{(3)}$ & 0.472 ± 0.132$^{(4)}$ & 0.916 ± 0.011$^{(3)}$ & 0.911 ± 0.010$^{(3)}$ & 445.061 ± 0.544$^{(5)}$ & 3448.023 ± 2.567$^{(4)}$ \\
CoVAE   & 0.473 ± 0.003$^{(4)}$ & \textbf{0.406 ± 0.002}$^{(1)}$ & 0.914 ± 0.008$^{(4)}$ & 0.860 ± 0.054$^{(5)}$ & \textbf{429.018 ± 0.515}$^{(1)}$ & 3442.293 ± 3.755$^{(2)}$ \\
\bottomrule
\end{tabular}
}
\caption{Comparison of multimodal VAE models on the pan-cancer mRNA/miRNA dataset on conditional tasks.}
\label{tab:conditional_results}
\end{table*}
What is of more interest is considering the performance of the various models in conditional tasks. Table\ref{tab:conditional_results} summarizes the performance of the models in conditional tasks, both in terms of Mean Absolute Error during the reconstruction of the missing modality and in the classification of the cancer type according to the missing modality\footnote{In this task, a classifier is trained on $z_k$ and the performance of the same classifier is evaluated on $z_{-k}$}. This time, joint and MoE models perform better, with CoVAE being the strongest performer on the reconstruction of mRNA from miRNA and one of the best models in the reverse reconstruction.

\begin{figure}[h]
    \centering
    \begin{subfigure}{\linewidth}
        \centering
        \includegraphics[width=0.8\linewidth]{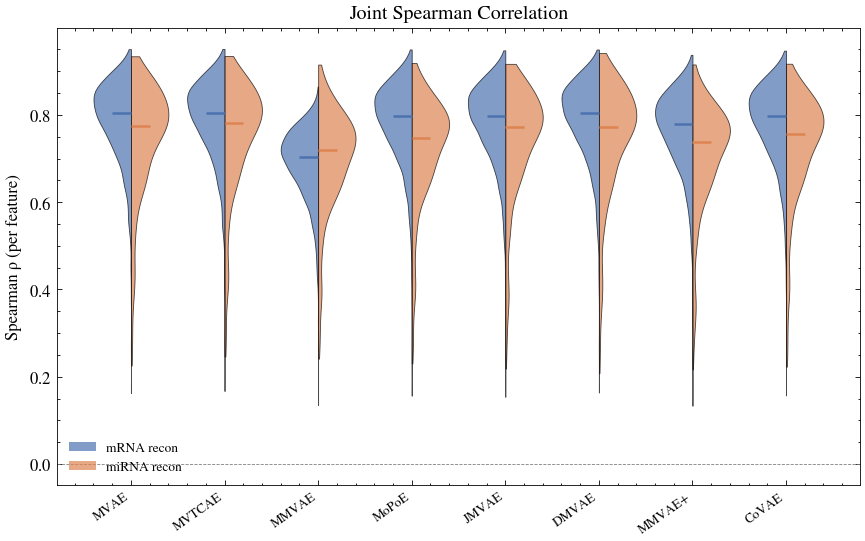}
    \end{subfigure}
    \begin{subfigure}{\linewidth}
        \centering
        \includegraphics[width=0.8\linewidth]{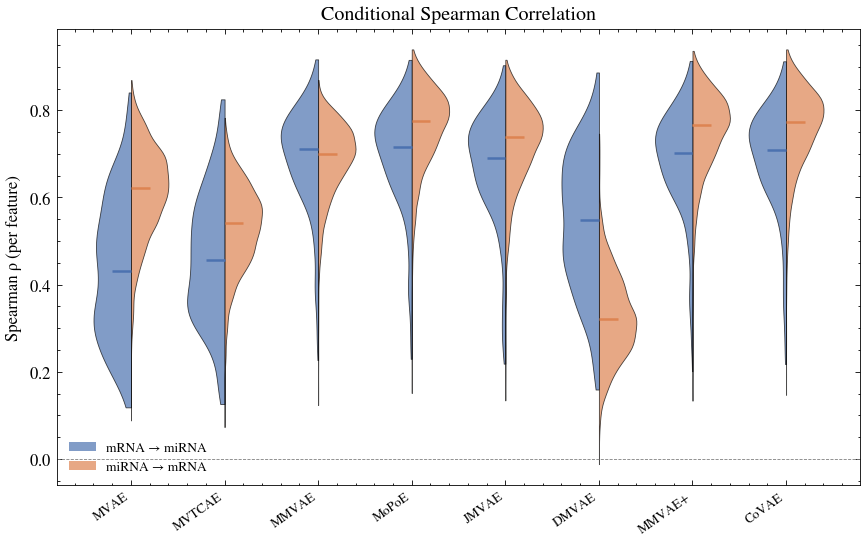}
    \end{subfigure}
    \caption{Violin plots for the distributions of the Spearman coefficients of the joint (top) and conditional (bottom) reconstructions for mRNA (blue) and miRNA (orange)}
    \label{fig:spearman}
\end{figure}

We further assess feature-level reconstruction fidelity in Fig.\ref{fig:spearman} (and Table\ref{tab:spearman}), showing the distribution of per-feature Spearman correlation coefficients between reconstructed and true values for both joint and conditional tasks. Among all tested models, only CoVAE, MoPoE, and JMVAE maintain consistently high correlations across all four settings (joint and conditional, for both modalities).

Overall, the MLOmics benchmark shows that CoVAE achieves competitive performance across the board while being the only model, alongside MoPoE and JMVAE, to avoid significant weaknesses in any of the evaluated settings. Importantly, it performs particularly well on the task that most aligns with the goal of the model, that is, generating unseen modalities with accurate statistics.

\section{Discussion} \label{sec:discussion}
In this work, we identified and proposed a solution to a common problem in multimodal integration through Variational Autoencoders: the collapse of the joint statistical structure induced by data fusion in the latent space. Although the consequences of this issue may not be relevant in some applications, such as those in which the main focus is only the projection of the data in a lower dimensional space, knowledge of the phenomenon is instead pivotal in  scenarios that require the accurate uncertainty quantification and conditional generation.

For this reason, we propose a class of models, CoVAE, that encodes the cross-modality correlations in the latent space. This approach provides an intuitive and motivated method to sample unobserved modalities in the latent space while circumventing the artificial correlations that other models introduce. We show through extensive testing on controlled simulated data that indeed CoVAE solves the problem, both in terms of accurately learning correlations between modalities and in terms of exploiting these correlations while quantifying uncertainty.
While the synthetic benchmarks highlight that CoVAE is the only model able to recover and reproduce the correct correlation, the MLOmics benchmark shows that it does so while achieving competitive results on downstream tasks typical of these models, and in particular it performs very well on the conditional generation task.

While promising, CoVAE has a number of limitations. First, the model assumes that any data correlation can be modeled as a global correlation in a Gaussian space, something that is unlikely in real world scenarios. Secondly, while the learned covariance structure helps the joint inference, the training process often cannot achieve the values of other models with the same values of latent dimension $D$ (see Fig. \ref{fig:nll_comparison}); we investigate in Appendix\ref{supp:jointNLL} how this gap in performance is not driven by the reconstruction quality but by the different geometry of the latent space. Moreover, as we see in Fig \ref{fig:example_conditional_generations}, CoVAE's predictive distributions in conditional generation are often broader when correlations are low: this is correct from a statistical perspective, but may lead to generating samples which are slightly outside the data manifold, thus appearing as out of focus digits in the example we considered. Such problems may be ameliorated or outright solved by assuming more complex prior structures or conditional prediction procedures, an issue that we plan to investigate in future work.

\begin{acknowledgements} The authors acknowledge support from the Italian Association for Cancer Research (AIRC) under grant IG 27631. GS acknowledges co-funding from Next Generation EU, in the context of the National Recovery and Resilience Plan, Investment PE1 - Project FAIR “Future Artificial Intelligence Research”. This resource was co-financed by the Next Generation EU [DM 1555 del 11.10.22].
\end{acknowledgements}

\bibliography{refs}

\newpage

\onecolumn

\title{CoVAE: correlated multimodal generative modeling\\(Supplementary Material)}
\maketitle

\appendix
\section{Normalizing Flows} \label{sec:normalizing_flows}
Normalizing Flows\cite{rezende2015variational} are a class of Neural Networks-based methods for constructing complex probability distributions starting from simple, tractable ones (often Gaussians). Defining $\mathbf{z}$ our latent variable, and $\bold{x}$ the observed data, belonging to an arbitrarily complex distribution, we decompose the invertible mapping $f:\mathbb{R}^{D} \rightarrow \mathbb{R}^{D}$ into a sequence of invertible transformations $\{f_{K} \}$ such that $$ \mathbf{x} = f_{k} \circ f_{k-1} \circ ... \circ f_{1}(\mathbf{z})$$
where every transformation is invertible. Propagating the signal from the data space $\mathbf{x}$ to the simple distribution $\mathbf{z}$ is referred to as \textit{forward} process, while the opposite is the \textit{inverse} process.
Every function $f_{k}$ is parameterized by a Neural Network, optimized by maximizing the likelihood of the data distribution $p_{X}(x)$

\begin{equation}
\log p_{X}(\mathbf{x})
  =
  \log p_{0}\bigl(\mathbf{z}\bigr)
  -
  \sum_{k=1}^{K}
      \log\Bigl|\det
            \Bigl(
              \frac{\partial f_{k}}{\partial \mathbf{h}_{k-1}}
            \Bigr)
          \Bigr|,
\end{equation}
$$
\mathbf{h}_{0}=\mathbf{z},
\;
\mathbf{h}_{k}=f_{k}(\mathbf{h}_{k-1}),
\;
\mathbf{x}=\mathbf{h}_{K}.
$$

A specific class of Normalizing Flows known for its fast backward process are the Masked Autoregressive Flows\cite{papamakarios2017masked} (MAF), in which each data dimension $x_i$ is generated autoregressively as
\begin{equation}
x_i = \mu_i(x_{1:i-1}) + \exp(\alpha_i(x_{1:i-1})) \, z_i, 
\qquad z_i \sim \mathcal{N}(0,1),
\end{equation}
where $\mu_i$ and $\alpha_i$ are outputs of masked neural networks ensuring the autoregressive dependency structure. 
This formulation defines an invertible mapping from latent noise $\mathbf{z}\sim \mathcal{N}(0,I)$ to data $\mathbf{x}$, with a triangular Jacobian whose log-determinant is
\begin{equation}
\log \left|\det \frac{\partial f^{-1}}{\partial \mathbf{x}}\right| = -\sum_{i=1}^{D} \alpha_i(x_{1:i-1}).
\end{equation}
\section{Experimental details} \label{supp:exp}

\subsection{Synthetic dataset}

\paragraph{Dataset generation} We create a dataset of $25000$ images following the procedure in Sec\ref{sec:experiments}. The transformed latent codes are decoded into images using a pretrained VAEGAN model (from the \texttt{pythae} library). The VAEGAN uses ResNet-based encoder and decoder architectures (\texttt{Encoder\_ResNet\_VAE\_MNIST}, \texttt{Decoder\_ResNet\_VAE\_MNIST}) with a convolutional discriminator (\texttt{Discriminator\_Conv\_MNIST}), and is trained for 50 epochs on MNIST with the AdamW optimizer (learning rate $10^{-4}$, batch size 64, adversarial loss scale 0.8, reconstruction layer 3, margin 0.4, equilibrium 0.68). The encoders and decoders of the models have been fixed and are reported in App.\ref{supp:architectures}. The resulting paired samples are randomly shuffled and split into 20,000 training and 5,000 test samples per configuration. Ground-truth cross-modal correlation is measured as the mean of the CCA components computed on the generative latent vectors.
\paragraph{Training}
All multimodal models are trained with the AdamW optimizer, a learning rate of $10^{-3}$, batch size 256, for 30 epochs, with a ReduceLROnPlateau learning rate scheduler (patience 30). The KL divergence weight is set to $\beta = 1$. A single random seed is used per configuration.

\paragraph{Representation extraction}
The representation of the data on which the CCA is evaluated is typically performed before the first linear layer by calling a pre-hook:
\begin{verbatim}
    self._h1 = _first_layer(self.model.decoders[mod_1]).register_forward_pre_hook(
    _pre(self._dec_in_1, "dec_mod_1")
)
self._h2 = _first_layer(self.model.decoders[mod_2]).register_forward_pre_hook(
    _pre(self._dec_in_2, "dec_mod_2")
)
\end{verbatim}

\paragraph{Use of normalizing flows}
For the plots shown in the main body of the paper, we choose to use a dataset created without the normalizing flows mapping the generated samples into the posteriors of the pretrained decoders. This is motivated by the preference of controlling exactly the correct latent linear correlation, which would be degraded by a non-linear normalizing flow. However, the qualitative result would be the same. Here we report the results for the 3 models that are able to generate uncorrelated data.

\begin{figure}[htbp]
    \centering
    \begin{tabular}{cc}
        \includegraphics[width=0.45\textwidth]{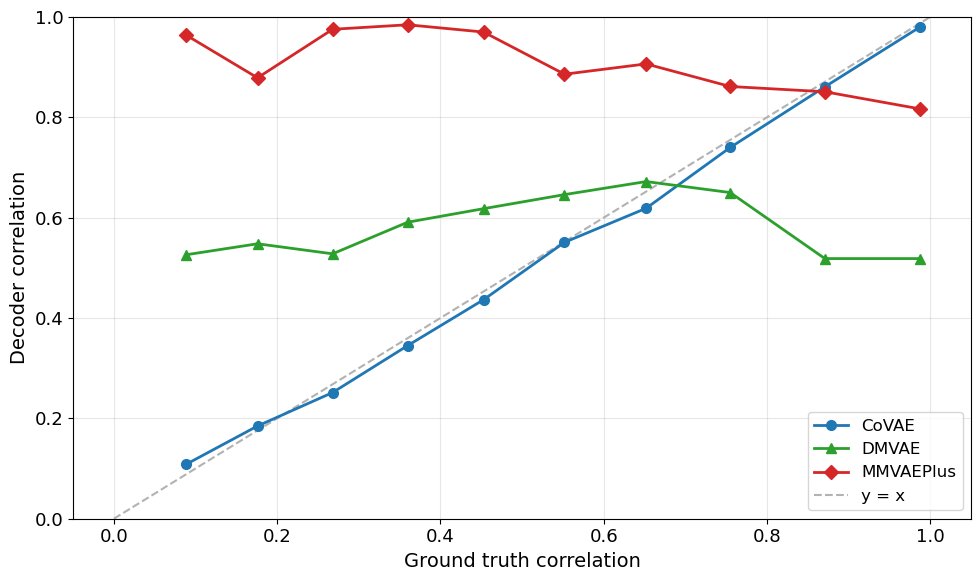} &
        \includegraphics[width=0.45\textwidth]{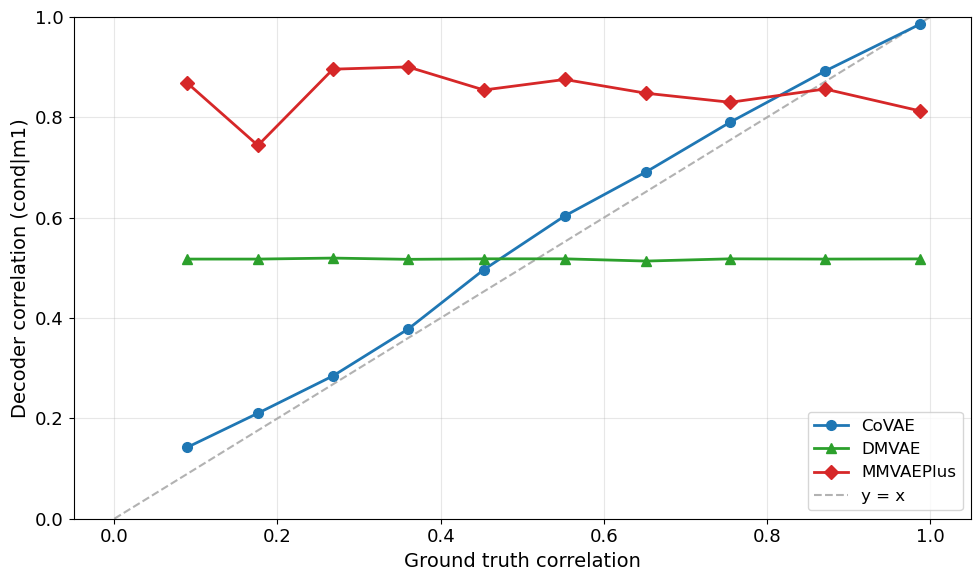} \\
        \includegraphics[width=0.45\textwidth]{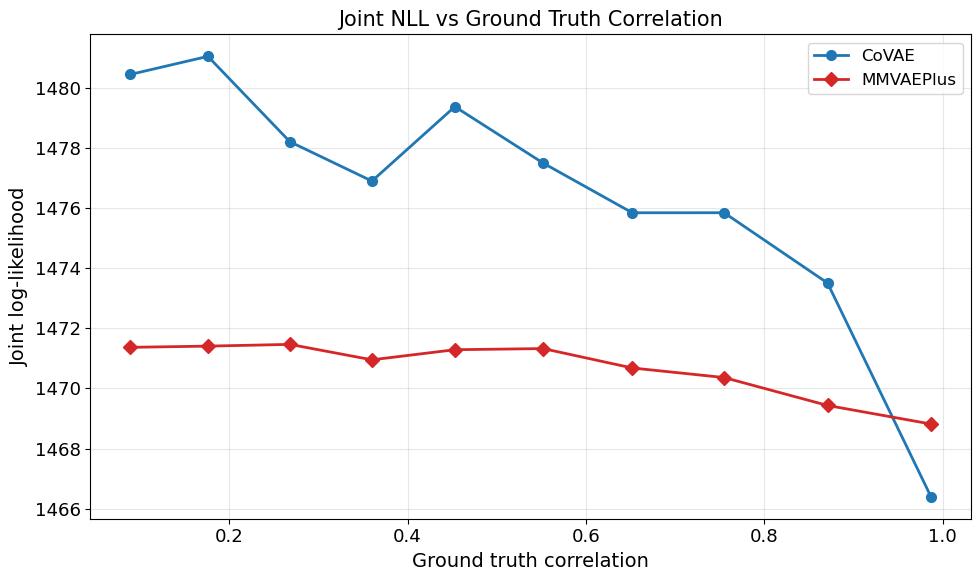} &
        \includegraphics[width=0.45\textwidth]{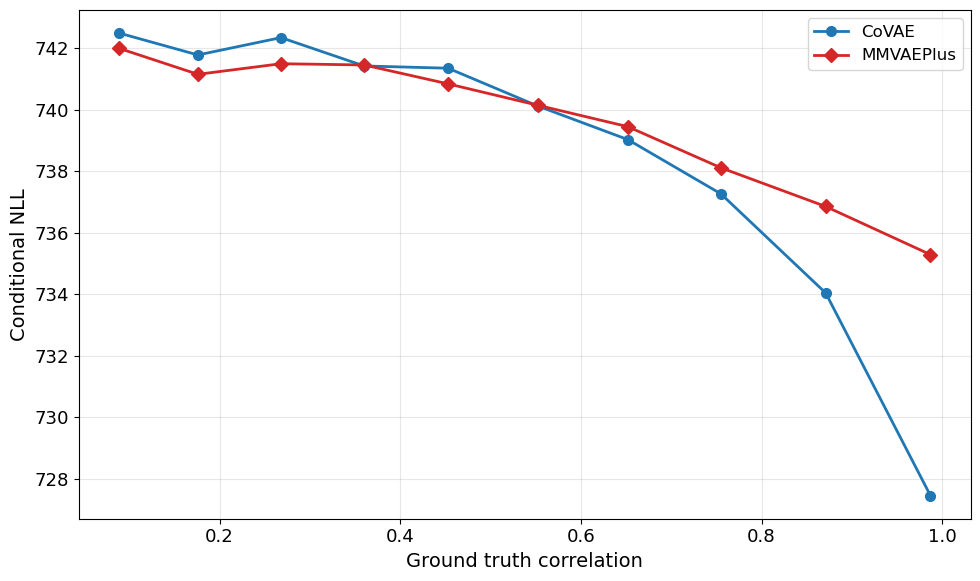} \\
    \end{tabular}
    \caption{Top left: joint correlation. Top right: conditional correlation. Bottom left: joint NLL. Bottom right: conditional NLL}
    \label{fig:flow_comparison}
\end{figure}

\paragraph{CoVAE options}

\begin{figure}[htbp]
    \centering
    \begin{tabular}{cc}
        \includegraphics[width=0.45\textwidth]{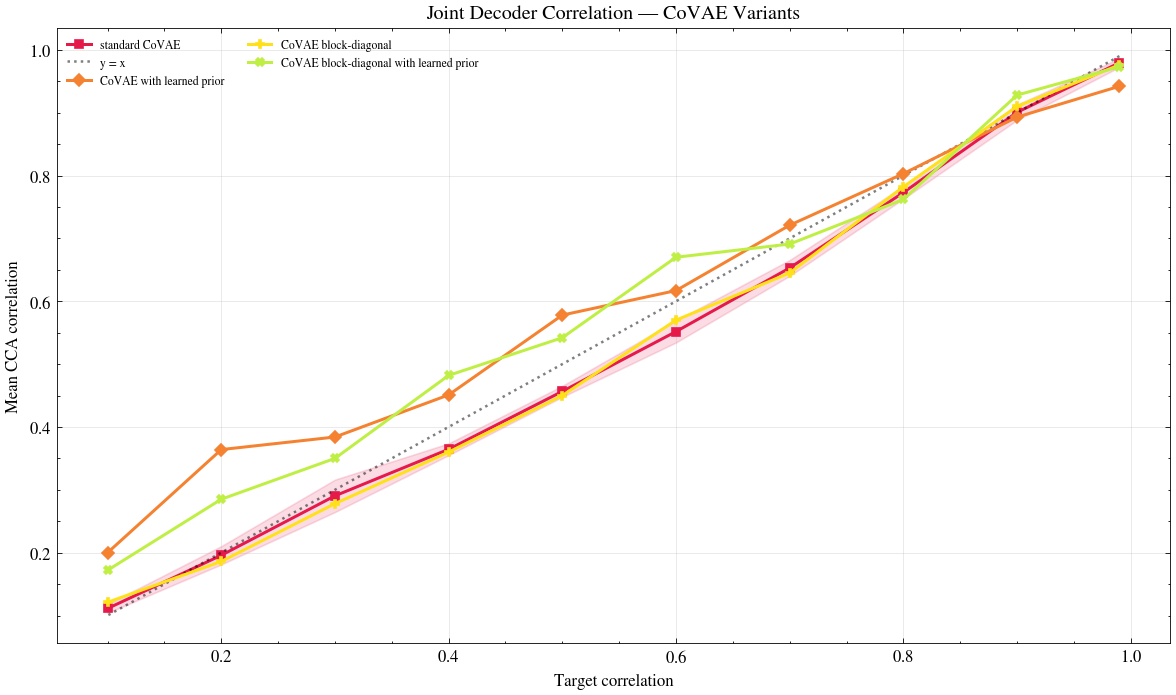} &
        \includegraphics[width=0.45\textwidth]{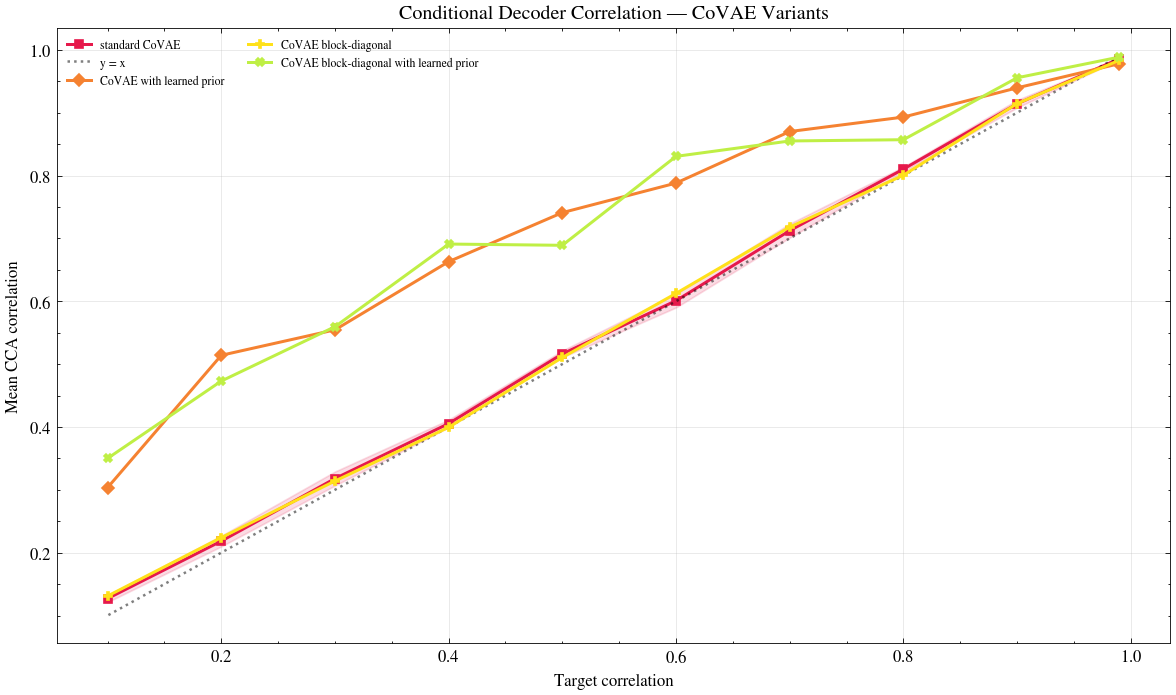} \\
        \includegraphics[width=0.45\textwidth]{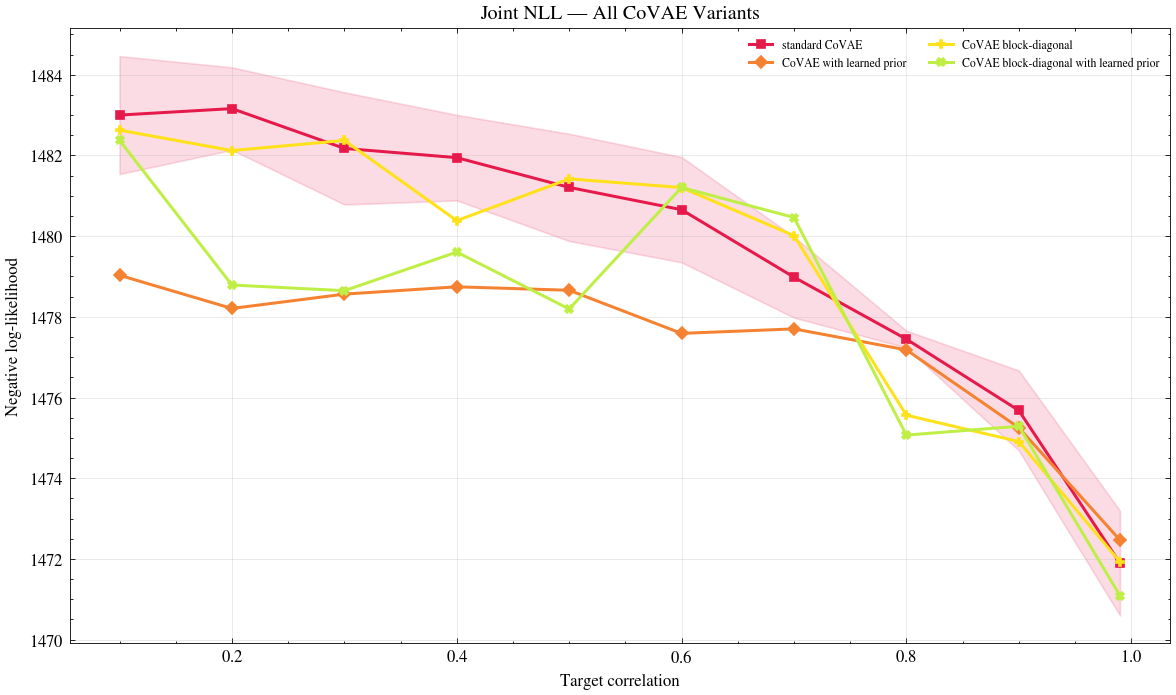} &
        \includegraphics[width=0.45\textwidth]{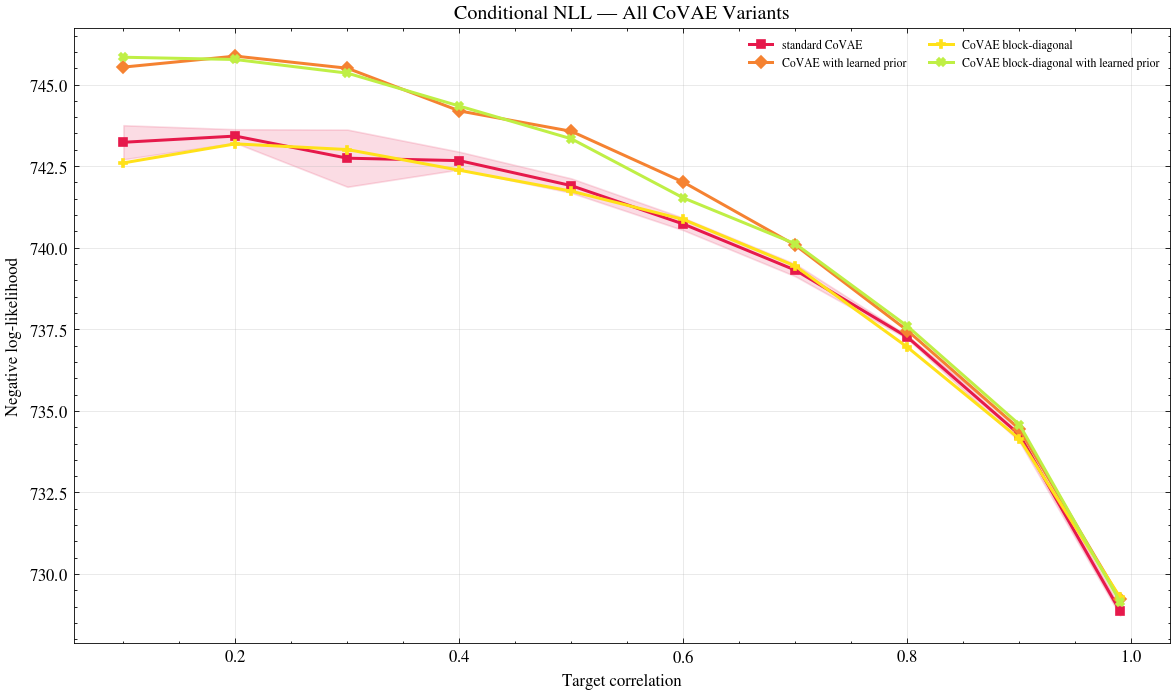} \\
    \end{tabular}
    \caption{Top left: joint correlation. Top right: conditional correlation. Bottom left: joint NLL. Bottom right: conditional NLL}
    \label{fig:covaes_comparison}
\end{figure}

Before arriving at the current defaults of CoVAE, we have tested many options. Fig.\ref{fig:covaes_comparison} shows that continuing to train the prior during the main training cycle degrades the performance, and that a different parameterization of the posterior, using cross diagonal blocks like \ref{eq:covariance_generation}, leads to no qualitative difference in performance. The reason for the degradation of the prior is that the model, trying to predict the missing modality during the cross-reconstruction step, overestimates the correlations in the latent space.

\subsection{MLOmics dataset}
\label{sec:mlomics_experiments}

We evaluate all models on the MLOmics Pan-cancer benchmark\citep{yang2025mlomics}\footnote{Dataset downloaded from \url{https://figshare.com/articles/dataset/MLOmics_Cancer_Multi-Omics_Database_for_Machine_Learning/28729127}}, which comprises 8{,}314 tumor samples across 32 cancer types with two transcriptomic modalities: mRNA gene expression (3{,}217 features) and miRNA expression (383 features). Features are standardized per-modality to zero mean and unit variance. The data is split into  train/test partitions with a proportion of 80/20, stratified by cancer type, and every experiment is repeated across five random seeds.

\paragraph{Metrics} 

We compare the ability of CoVAE to perform cancer type classification against the aforementioned models. For each of them, we measure the following metrics:
\begin{itemize}
    \item PREC (accuracy): $\text{PREC} = \frac{1}{N}\sum_{i=1}^{N} \mathbf{1}[\hat{y}_i = y_i]$
    \item NMI (normalized mutual information): $\text{NMI}(\hat{Y}, Y) = \frac{2, I(\hat{Y}; Y)}{H(\hat{Y}) + H(Y)}$ where $I$ is mutual information and $H$ is entropy.
    \item ARI (adjusted Rand index): $\text{ARI} = \frac{\text{RI} - \mathbb{E}[\text{RI}]}{\max(\text{RI}) - \mathbb{E}[\text{RI}]}$ where $\text{RI} = \frac{a + b}{\binom{N}{2}}$, with $a$ = pairs correctly grouped together, $b$ = pairs correctly separated.
    \item Spearman correlation: $r_s = 1 - \frac{6\sum_{i=1}^{N}d_i^2}{N(N^2-1)}$, where $d_i = \mathrm{rk}(x_i) - \mathrm{rk}(y_i)$ is the difference between the ranks of the predicted and true values, measuring the monotonic dependence between them, ranging from $-1$ (perfect negative) to $1$ (perfect positive).
\end{itemize}

\begin{figure}[h]
    \centering
    \includegraphics[width=0.7\linewidth]{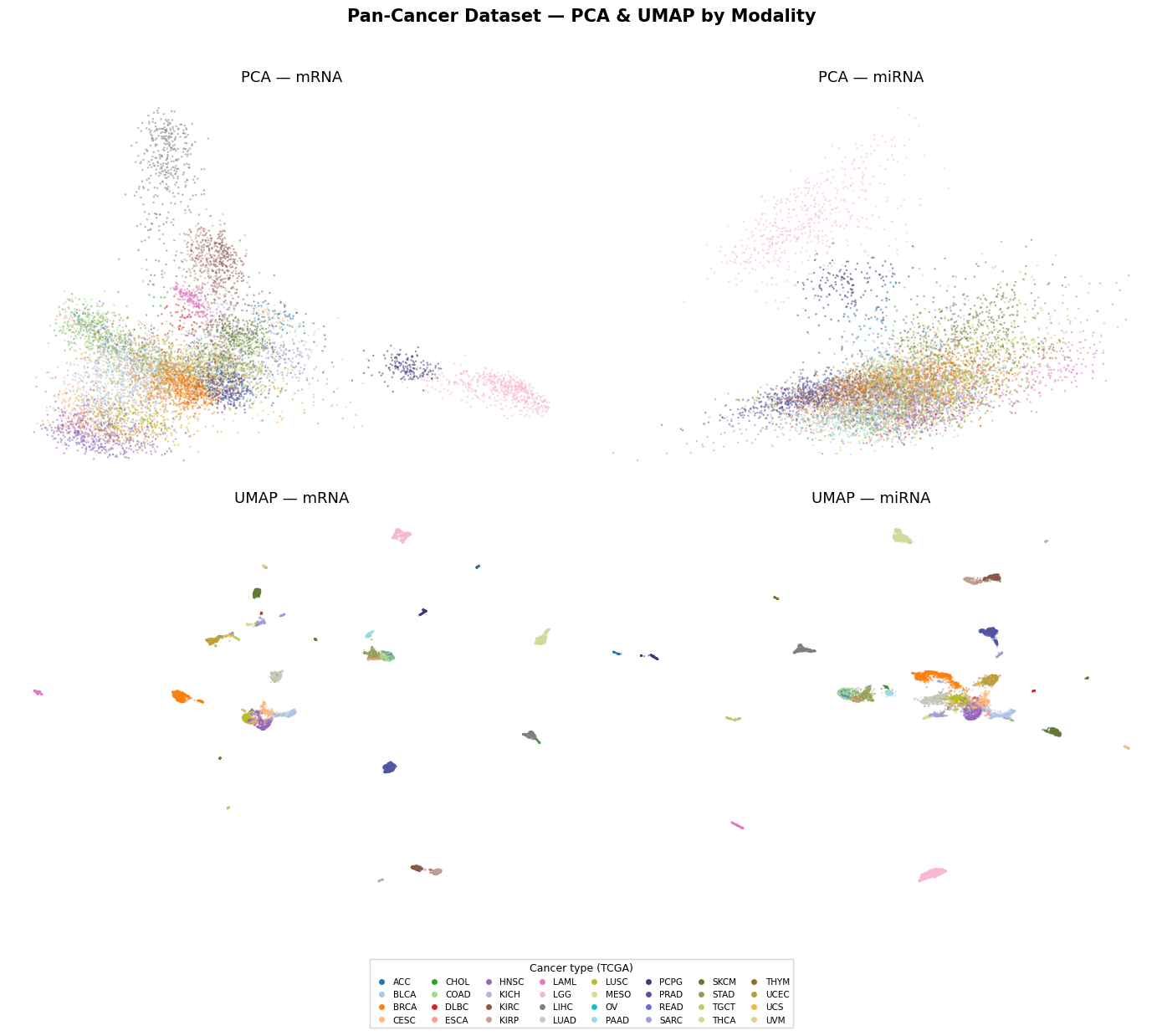}
    \caption{PCA and UMAP visualization of the dataset according to the two modalities}
    \label{fig:mlomics_umap}
\end{figure}

\begin{figure}[h]
    \centering
    \includegraphics[width=0.7\linewidth]{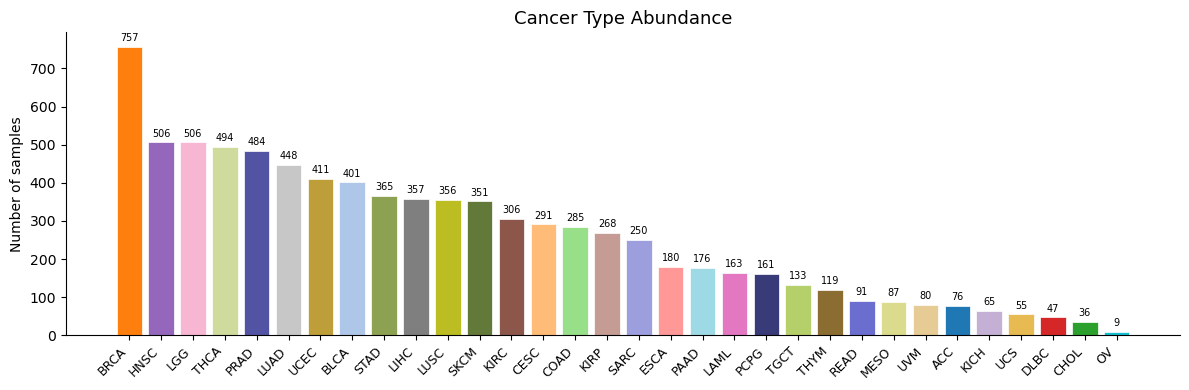}
    \caption{Labels distribution in the MLOmics pancancer dataset}
    \label{fig:mlomics_labels}
\end{figure}

\paragraph{Models.}
We compare the  baseline multimodal VAEs against three CoVAE configurations. All CoVAE variants use full-covariance per-sample KL regularization with a Cholesky-parameterized posterior and distance correlation (dcor) for pre-training the prior correlation structure (50 DCCA epochs, learning rate $10^{-3}$). They differ in how the 32 latent dimensions are allocated between modalities and whether marginal variance is learned.

\paragraph{Training.}
All models are trained for 200 epochs with AdamW (learning rate $10^{-3}$, batch size 128) and a ReduceLROnPlateau scheduler (patience 30). 

\begin{table}[h]
\centering
\begin{tabular}{lcccc}
\toprule
 & mRNA→miRNA & miRNA→mRNA & Joint mRNA & Joint miRNA \\
Model &  &  &  &  \\
\midrule
MVAE & 0.432 +/- 0.055 & 0.578 +/- 0.061 & \textbf{0.788 +/- 0.002} & 0.742 +/- 0.002 \\
MVTCAE & 0.465 +/- 0.014 & 0.547 +/- 0.028 & 0.787 +/- 0.001 & \textbf{0.749 +/- 0.002} \\
MMVAE & 0.685 +/- 0.004 & 0.705 +/- 0.012 & 0.704 +/- 0.009 & 0.695 +/- 0.004 \\
MoPoE & \textbf{0.691 +/- 0.002} & \textbf{0.758 +/- 0.002} & 0.781 +/- 0.001 & 0.716 +/- 0.001 \\
JMVAE & 0.668 +/- 0.002 & 0.723 +/- 0.001 & 0.782 +/- 0.001 & 0.735 +/- 0.001 \\
DMVAE & 0.531 +/- 0.019 & 0.284 +/- 0.026 & 0.788 +/- 0.001 & 0.740 +/- 0.002 \\
MMVAE+ & 0.682 +/- 0.003 & 0.736 +/- 0.031 & 0.758 +/- 0.009 & 0.701 +/- 0.011 \\
CoVAE & 0.684 +/- 0.002 & 0.757 +/- 0.002 & 0.779 +/- 0.002 & 0.723 +/- 0.002 \\
\bottomrule
\end{tabular}
\caption{Table of Spearman coefficients represented in Fig.\ref{fig:spearman}}
\label{tab:spearman}
\end{table}

\section{Joint NLL} \label{supp:jointNLL}

The joint NLL is significantly higher for CoVAE compared to the other architectures. Considering that a qualitative inspection does not reveal any noticeable drop in performance, we investigate further the reason behind this discrepancy.

\paragraph{Joint NLL decomposition}

We recall that the IWAE estimator is a tighter lower bound on the evidence compared to the standard ELBO, as instead of drawing a single sample from the approximate posterior, it draws $K$ samples for each data point and takes a weighted average of the likelihood:
$$\mathcal{L}_{\text{IWAE}}^K = \mathbb{E}_{z_1, \ldots, z_K \sim q_\phi(z|x)} \left[ \log \frac{1}{K} \sum_{k=1}^{K} \frac{p_\theta(x, z_k)}{q_\phi(z_k|x)} \right]$$

Equivalently, we can decompose it as a sum of 3 different terms:

\begin{equation} \label{eq:iwae_decomposition}
\mathcal{L}_{\text{IWAE}}^K = \mathbb{E}_{z_1, \ldots, z_K \sim q_\phi(z|x)} \Bigg[ \log \frac{1}{K} \sum_{k=1}^{K} \exp\!\Big( \underbrace{\log p_\theta(x|z_k)}_{\text{reconstruction}} + \underbrace{\log p(z_k)}_{\text{prior}} - \underbrace{\log q_\phi(z_k|x)}_{\text{posterior}} \Big) \Bigg]
\end{equation}

\begin{figure}[h]
    \centering
    \includegraphics[width=0.8\columnwidth]{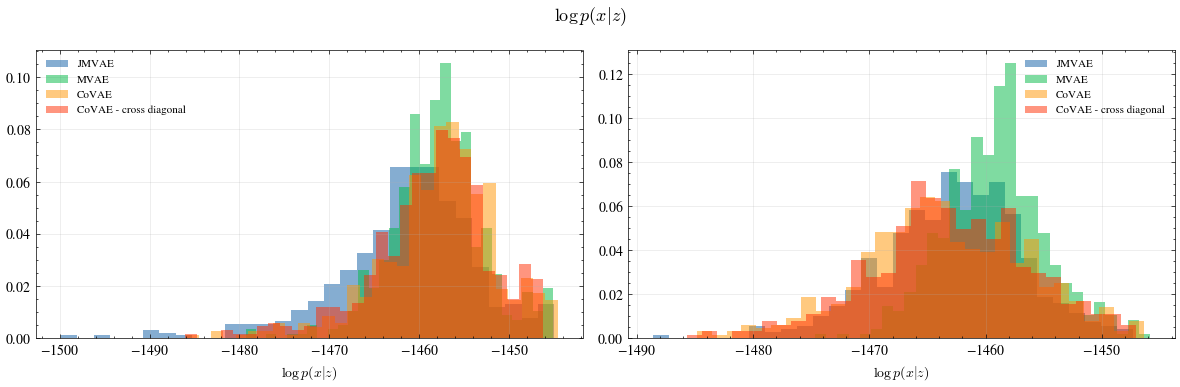}\\[0.5em]
    \includegraphics[width=0.8\columnwidth]{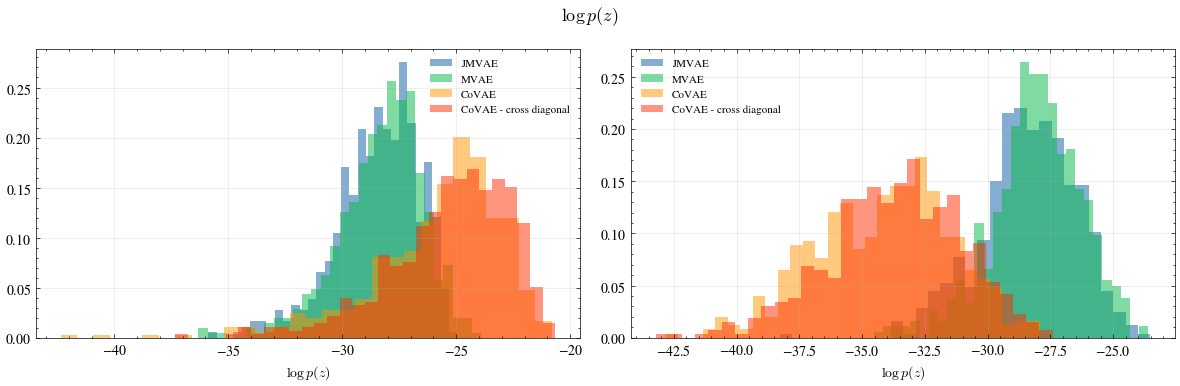}\\[0.5em]
    \includegraphics[width=0.8\columnwidth]{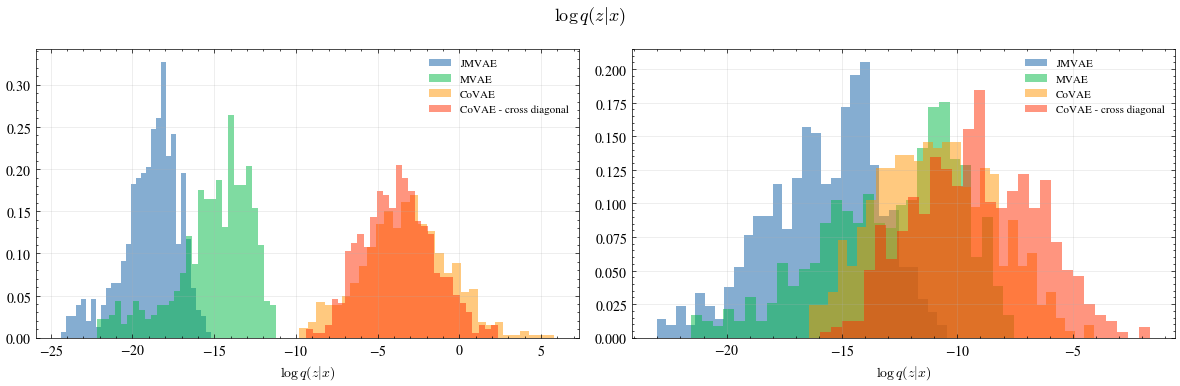}
    \caption{Decomposition of the terms in Eq.\ref{eq:iwae_decomposition} for JMVAE (blue), MVAE (green), CoVAE with the Cholesky covariance parameterization(yellow) and cross-diagonal (red). For each model, we sample 5000 times from the posteriors of 500 points from the test samples. Dataset correlation is $\rho=0.9$ (left) and $\rho=0.3$ (right)}
    \label{fig:iwae_decomposition}
\end{figure}

In Fig.\ref{fig:iwae_decomposition} we decompose the joint NLL estimation into its components and compare the distributions with JMVAE and MVAE at both high ($\rho=0.9$) and low ($\rho=0.3$) correlation. Furthermore, we test if the problem lies in the parameterization of the covariance through the Cholesky decomposition by adding the comparison with a variant of CoVAE in which the posterior is parameterized using a cross-diagonal structure equivalent to \ref{eq:covariance_generation}; in this case, the off-diagonal parameters are optimized directly and bounded by a sigmoid to be between 0 and 1 in order to guarantee positive definiteness\footnote{This method only guarantees positive definiteness with two modalities}. We notice that the likelihood term is comparable in all three models, and the difference is driven by the prior and posterior terms.

In particular, we observe that the prior term is higher in the case of high correlation and interestingly worse at lower correlation, while the posterior term contributes negatively at high correlation while being comparable at low correlation. Furthermore, we observe that the parameterization of the posterior does not lead to a significant difference in distributions.
\begin{figure}[h]
    \centering
    \includegraphics[width=0.8\columnwidth]{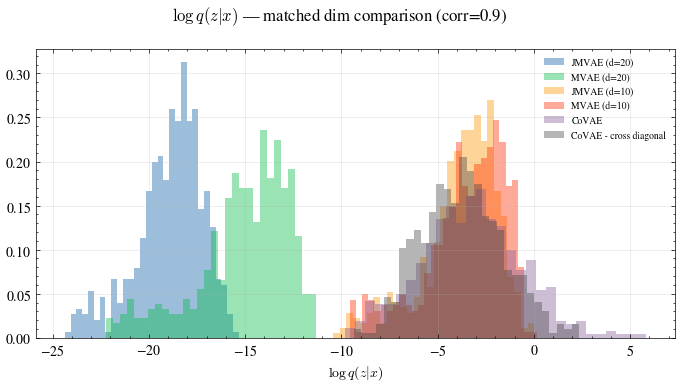}\\[0.5em]
    \caption{Comparison of CoVAE models' posterior density with the one of JMVAE and MVAE with two different latent dimensions}
    \label{fig:iwae_decomposition_explanation}
\end{figure}

However, this discrepancy can be explained by observing that the correlated prior in practice forces the data to lie on a lower-dimensional manifold. To test this hypothesis, we train JMVAE and MVAE on the dataset $\rho=0.9$ while setting $d(z)=10$. Fig.~\ref{fig:iwae_decomposition_explanation} shows a very good overlap in the entropy term, highlighting how the discrepancy in the Joint NLL can be explained by the correlated parameterization of the prior, which effectively embeds a lower-dimensional manifold in the higher-dimensional latent space. This is further confirmed by computing the participation ratio $d_{\text{eff}} = \left(\sum_i \lambda_i\right)^2 / \sum_i \lambda_i^2$ from the eigenvalues of the prior covariance matrix, which yields an effective dimensionality close to $11$, consistent with the latent dimension used in the control experiment.
\section{Architectures} \label{supp:architectures}

We report the architectures of encoders and decoders used in all the tests and standardized across the models.

\paragraph{Synthetic data}

\begin{verbatim}
    
Encoder_ResNet_VAE_MNIST(
  (layers): ModuleList(
    (0): Sequential(
      (0): Conv2d(1, 64, kernel_size=(4, 4), stride=(2, 2), padding=(1, 1))
    )
    (1): Sequential(
      (0): Conv2d(64, 128, kernel_size=(4, 4), stride=(2, 2), padding=(1, 1))
    )
    (2): Sequential(
      (0): Conv2d(128, 128, kernel_size=(3, 3), stride=(2, 2), padding=(1, 1))
    )
    (3): Sequential(
      (0): ResBlock(
        (conv_block): Sequential(
          (0): ReLU()
          (1): Conv2d(128, 32, kernel_size=(3, 3), stride=(1, 1), padding=(1, 1))
          (2): ReLU()
          (3): Conv2d(32, 128, kernel_size=(1, 1), stride=(1, 1))
        )
      )
      (1): ResBlock(
        (conv_block): Sequential(
          (0): ReLU()
          (1): Conv2d(128, 32, kernel_size=(3, 3), stride=(1, 1), padding=(1, 1))
          (2): ReLU()
          (3): Conv2d(32, 128, kernel_size=(1, 1), stride=(1, 1))
        )
      )
    )
  )
  (embedding): Linear(in_features=2048, out_features=20, bias=True)
  (log_var): Linear(in_features=2048, out_features=20, bias=True)
)

--- Standard Decoder (Decoder_ResNet_AE_MNIST) ---
    latent_dim=20, input_dim=(1,28,28)
Decoder_ResNet_AE_MNIST(
  (layers): ModuleList(
    (0): Linear(in_features=20, out_features=2048, bias=True)
    (1): ConvTranspose2d(128, 128, kernel_size=(3, 3), stride=(2, 2), padding=(1, 1))
    (2): Sequential(
      (0): ResBlock(
        (conv_block): Sequential(
          (0): ReLU()
          (1): Conv2d(128, 32, kernel_size=(3, 3), stride=(1, 1), padding=(1, 1))
          (2): ReLU()
          (3): Conv2d(32, 128, kernel_size=(1, 1), stride=(1, 1))
        )
      )
      (1): ResBlock(
        (conv_block): Sequential(
          (0): ReLU()
          (1): Conv2d(128, 32, kernel_size=(3, 3), stride=(1, 1), padding=(1, 1))
          (2): ReLU()
          (3): Conv2d(32, 128, kernel_size=(1, 1), stride=(1, 1))
        )
      )
      (2): ReLU()
    )
    (3): Sequential(
      (0): ConvTranspose2d(128, 64, kernel_size=(3, 3), stride=(2, 2), padding=(1, 1), output_padding=(1, 1))
      (1): ReLU()
    )
    (4): Sequential(
      (0): ConvTranspose2d(64, 1, kernel_size=(3, 3), stride=(2, 2), padding=(1, 1), output_padding=(1, 1))
      (1): Sigmoid()
    )
  )
)

\end{verbatim}

\paragraph{MLOmics}
\begin{verbatim}
======================================================================
  MLOMICS (mRNA + miRNA)
======================================================================

--- mRNA Encoder (OmicsEncoder) ---
    input_dim=3217, latent_dim=32, hidden_dims=(512,256,128), dropout=0.1
OmicsEncoder(
  (encoder): Sequential(
    (0): Linear(in_features=3217, out_features=512, bias=True)
    (1): BatchNorm1d(512, eps=1e-05, momentum=0.1, affine=True, track_running_stats=True)
    (2): ReLU()
    (3): Dropout(p=0.1, inplace=False)
    (4): Linear(in_features=512, out_features=256, bias=True)
    (5): BatchNorm1d(256, eps=1e-05, momentum=0.1, affine=True, track_running_stats=True)
    (6): ReLU()
    (7): Dropout(p=0.1, inplace=False)
    (8): Linear(in_features=256, out_features=128, bias=True)
    (9): BatchNorm1d(128, eps=1e-05, momentum=0.1, affine=True, track_running_stats=True)
    (10): ReLU()
    (11): Dropout(p=0.1, inplace=False)
  )
  (mu): Linear(in_features=128, out_features=32, bias=True)
  (logvar): Linear(in_features=128, out_features=32, bias=True)
)

--- mRNA Decoder (OmicsDecoder) ---
    latent_dim=32, output_dim=3217, hidden_dims=(128,256,512), dropout=0.1
OmicsDecoder(
  (decoder): Sequential(
    (0): Linear(in_features=32, out_features=128, bias=True)
    (1): BatchNorm1d(128, eps=1e-05, momentum=0.1, affine=True, track_running_stats=True)
    (2): ReLU()
    (3): Dropout(p=0.1, inplace=False)
    (4): Linear(in_features=128, out_features=256, bias=True)
    (5): BatchNorm1d(256, eps=1e-05, momentum=0.1, affine=True, track_running_stats=True)
    (6): ReLU()
    (7): Dropout(p=0.1, inplace=False)
    (8): Linear(in_features=256, out_features=512, bias=True)
    (9): BatchNorm1d(512, eps=1e-05, momentum=0.1, affine=True, track_running_stats=True)
    (10): ReLU()
    (11): Dropout(p=0.1, inplace=False)
    (12): Linear(in_features=512, out_features=3217, bias=True)
  )
)

--- miRNA Encoder (OmicsEncoder) ---
    input_dim=383, latent_dim=32, hidden_dims=(128,64), dropout=0.1
OmicsEncoder(
  (encoder): Sequential(
    (0): Linear(in_features=383, out_features=128, bias=True)
    (1): BatchNorm1d(128, eps=1e-05, momentum=0.1, affine=True, track_running_stats=True)
    (2): ReLU()
    (3): Dropout(p=0.1, inplace=False)
    (4): Linear(in_features=128, out_features=64, bias=True)
    (5): BatchNorm1d(64, eps=1e-05, momentum=0.1, affine=True, track_running_stats=True)
    (6): ReLU()
    (7): Dropout(p=0.1, inplace=False)
  )
  (mu): Linear(in_features=64, out_features=32, bias=True)
  (logvar): Linear(in_features=64, out_features=32, bias=True)
)

--- miRNA Decoder (OmicsDecoder) ---
    latent_dim=32, output_dim=383, hidden_dims=(64,128), dropout=0.1
OmicsDecoder(
  (decoder): Sequential(
    (0): Linear(in_features=32, out_features=64, bias=True)
    (1): BatchNorm1d(64, eps=1e-05, momentum=0.1, affine=True, track_running_stats=True)
    (2): ReLU()
    (3): Dropout(p=0.1, inplace=False)
    (4): Linear(in_features=64, out_features=128, bias=True)
    (5): BatchNorm1d(128, eps=1e-05, momentum=0.1, affine=True, track_running_stats=True)
    (6): ReLU()
    (7): Dropout(p=0.1, inplace=False)
    (8): Linear(in_features=128, out_features=383, bias=True)
  )
)
\end{verbatim}
\section{Example Reconstructions}

\begin{figure}[h]
    \centering
    \begin{subfigure}{0.40\textwidth}
        \centering
        \includegraphics[width=\textwidth]{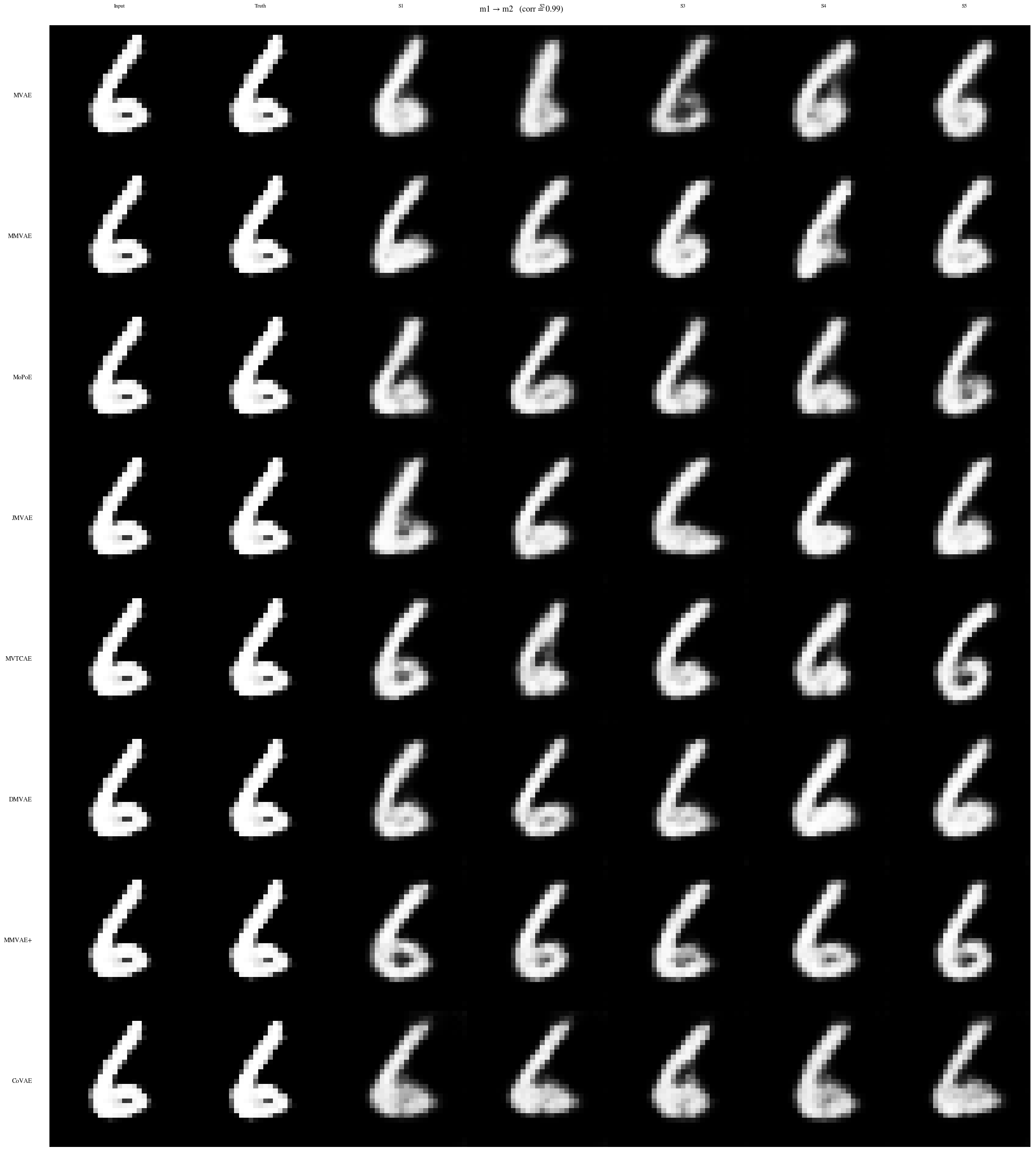}
    \end{subfigure}
    \hfill
    \begin{subfigure}{0.40\textwidth}
        \centering
        \includegraphics[width=\textwidth]{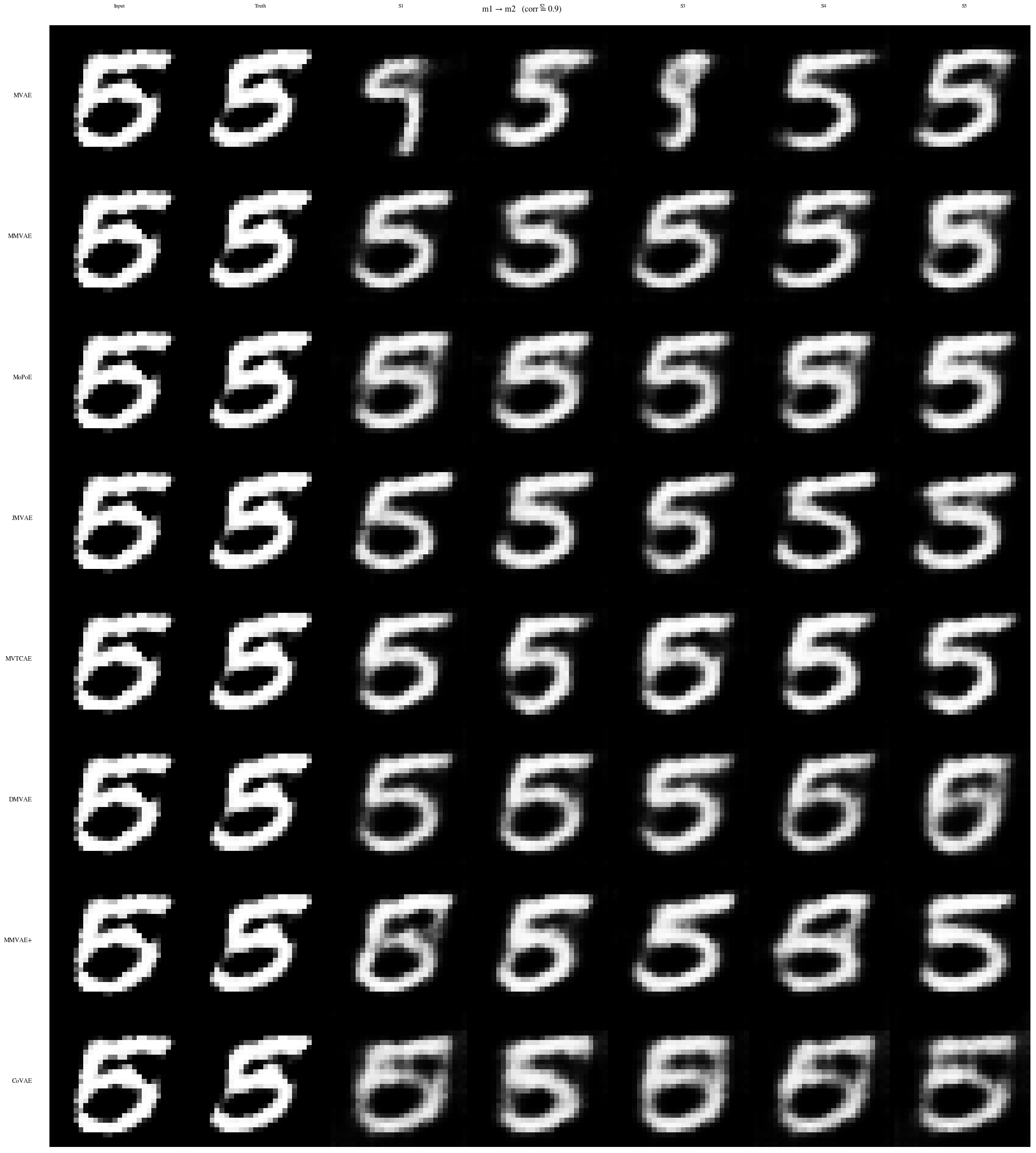}
    \end{subfigure}
    \vspace{0.3em}
    \begin{subfigure}{0.40\textwidth}
        \centering
        \includegraphics[width=\textwidth]{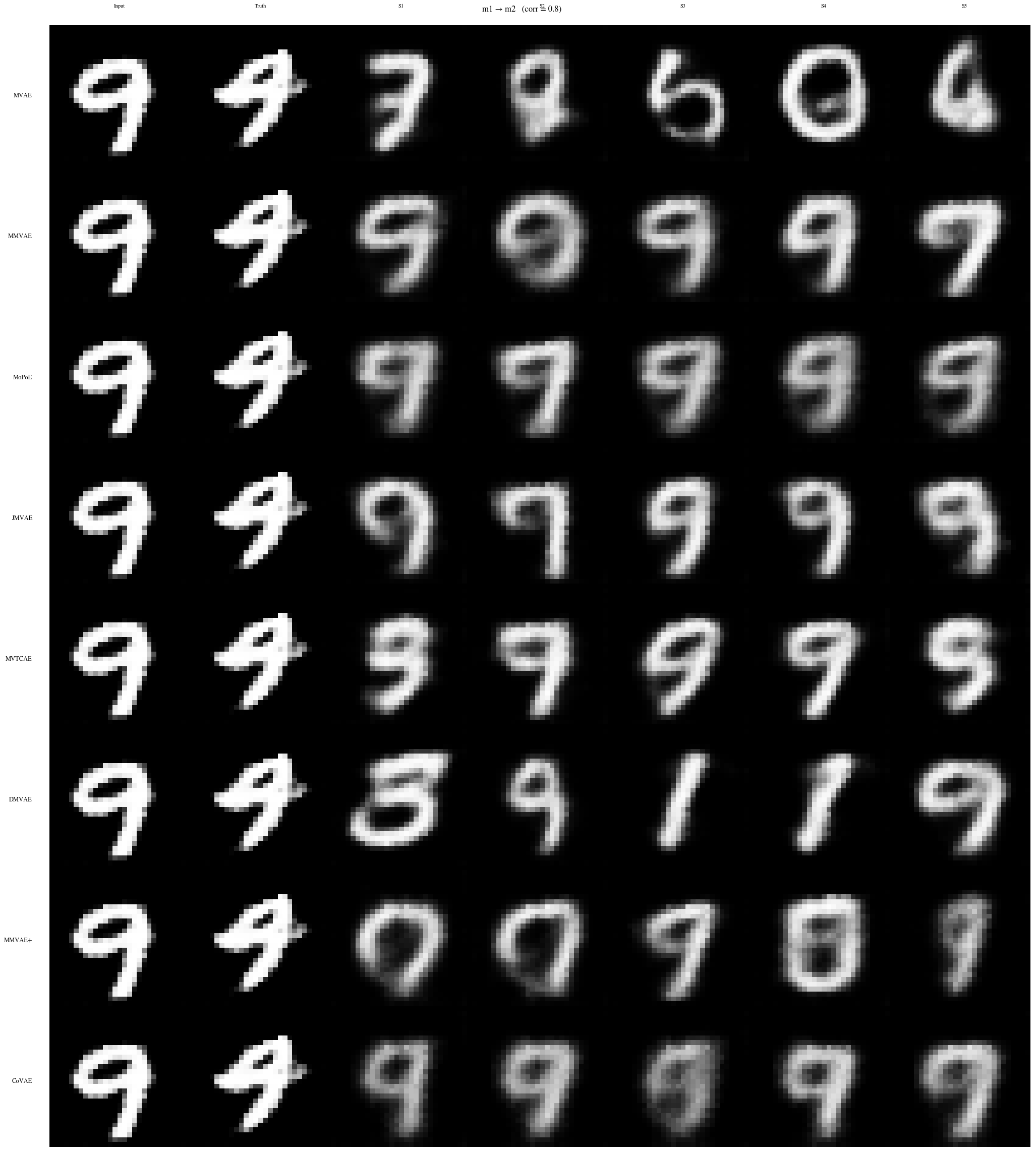}
    \end{subfigure}
    \hfill
    \begin{subfigure}{0.40\textwidth}
        \centering
        \includegraphics[width=\textwidth]{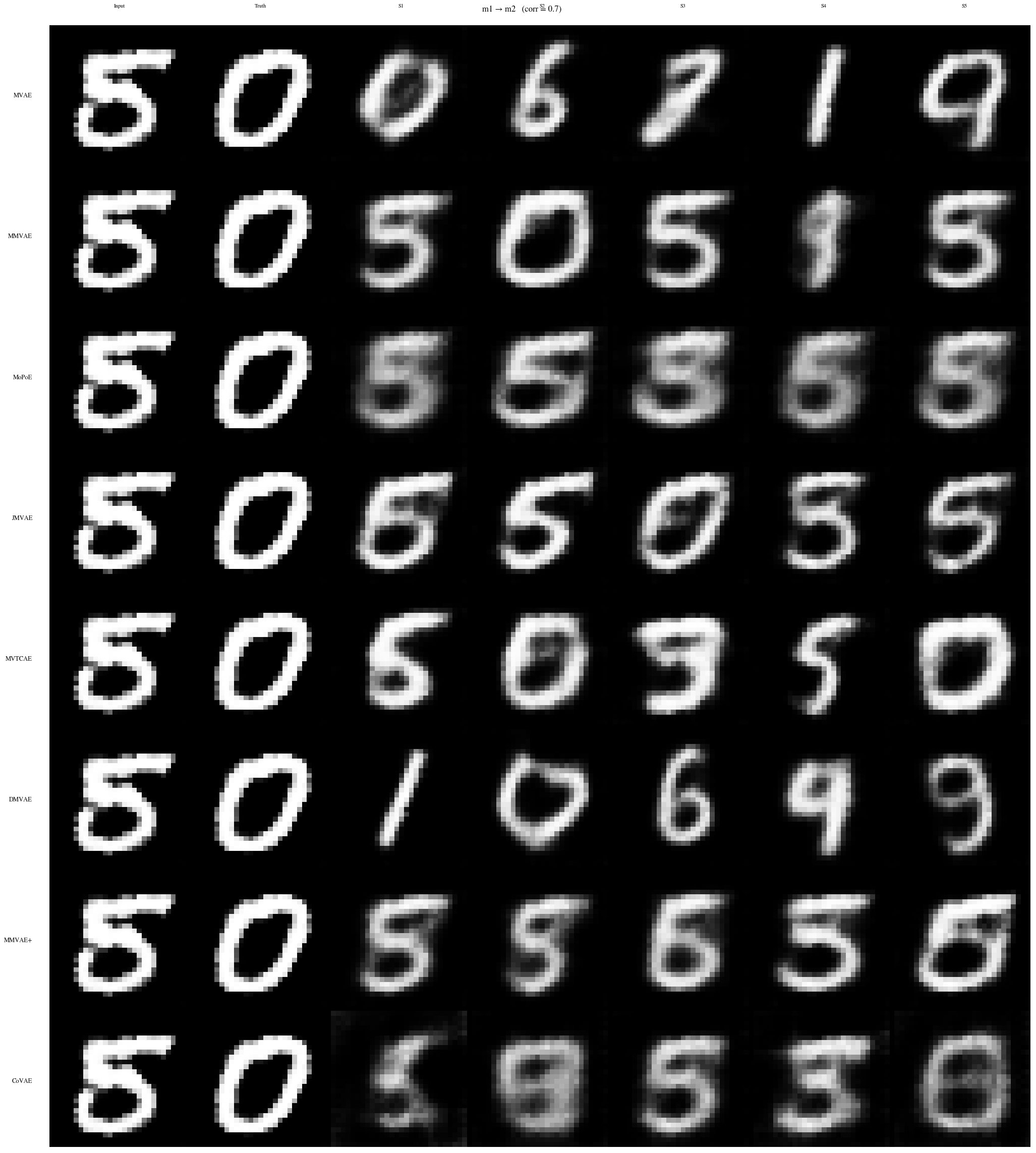}
    \end{subfigure}
    \vspace{0.3em}
    \begin{subfigure}{0.40\textwidth}
        \centering
        \includegraphics[width=\textwidth]{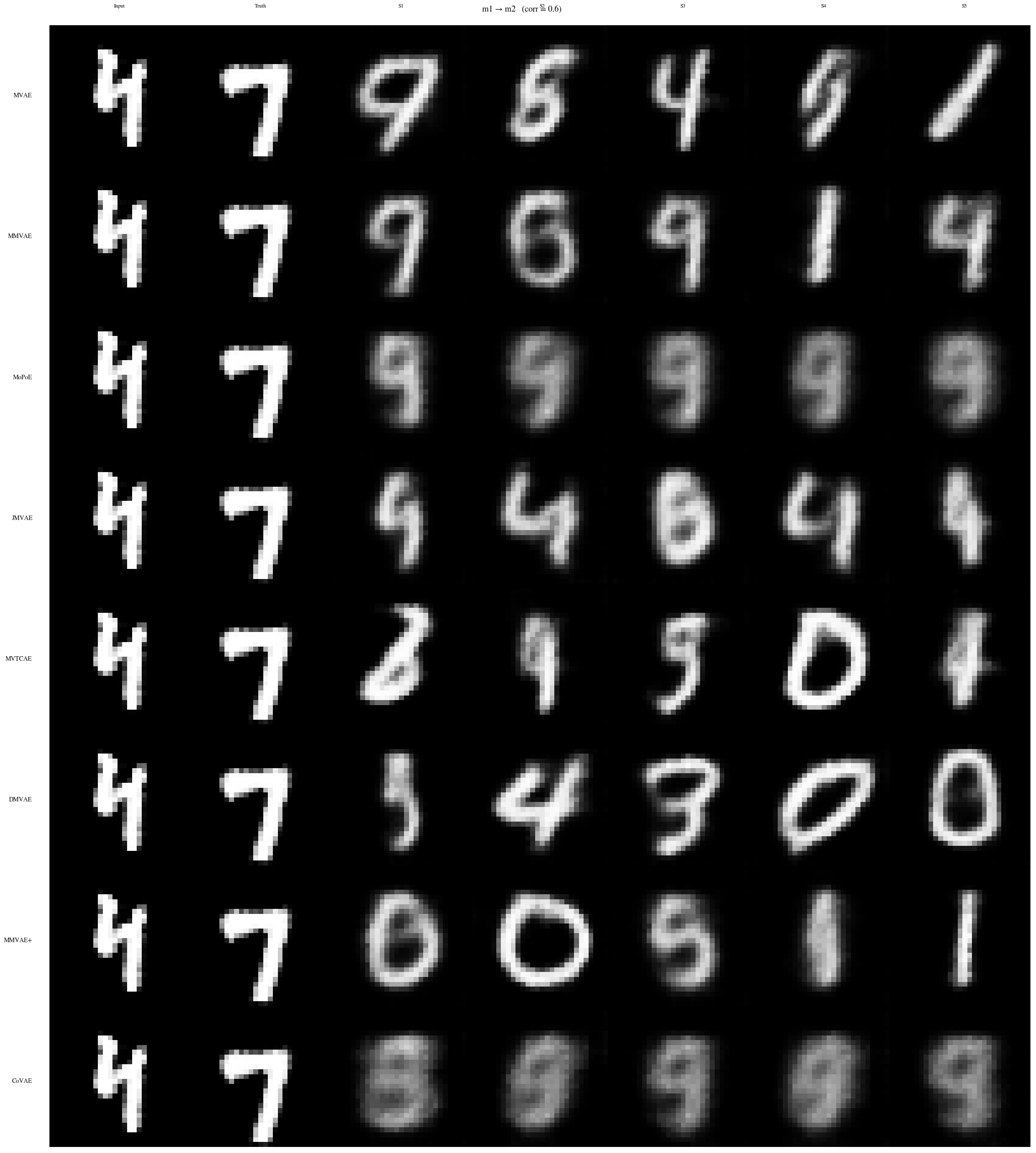}
    \end{subfigure}
    \hfill
    \begin{subfigure}{0.40\textwidth}
        \centering
        \includegraphics[width=\textwidth]{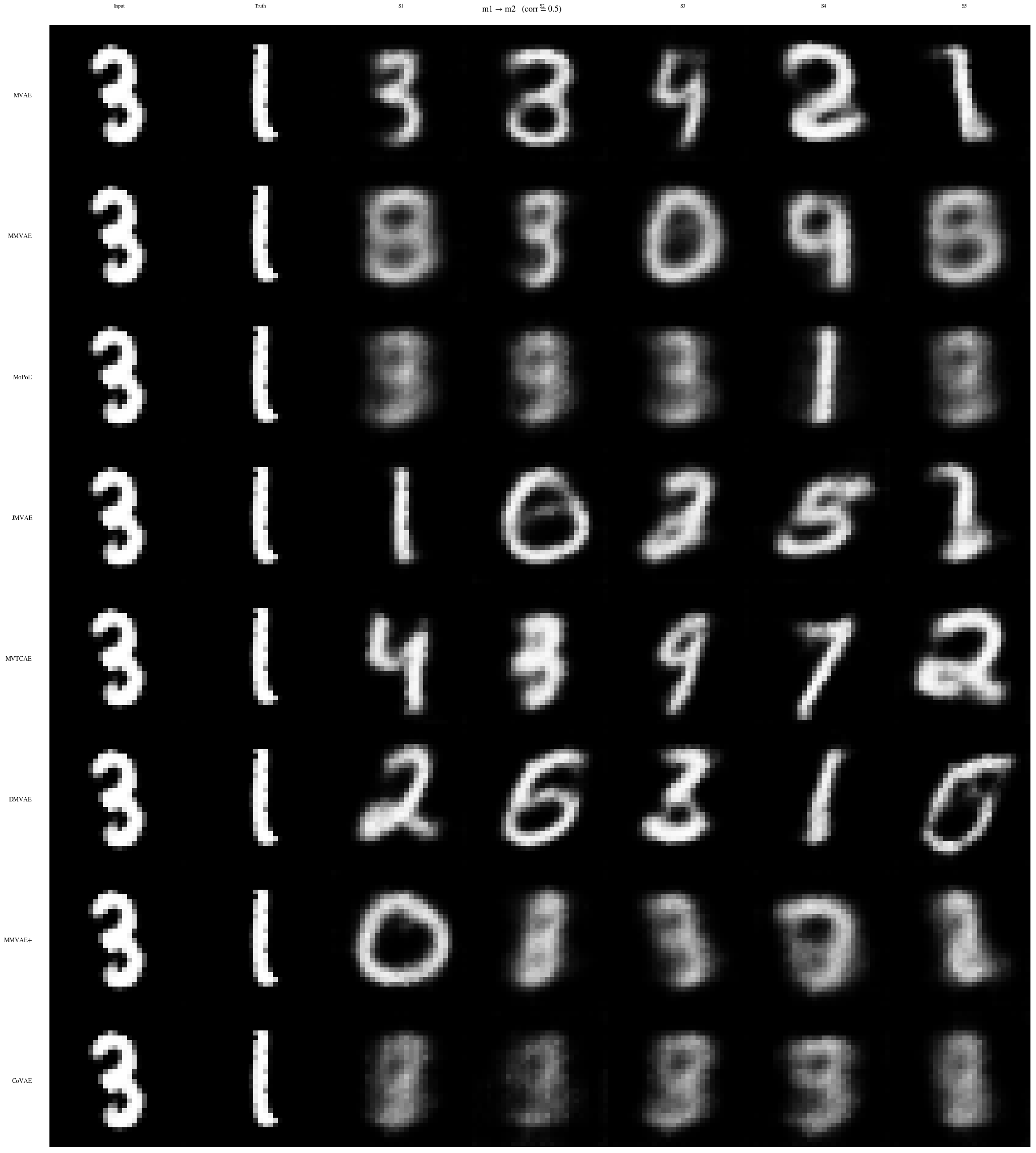}
    \end{subfigure}
    \caption{Conditional reconstructions for all models tested at different correlation levels}
    \label{fig:examples_conditional}
\end{figure}

\paragraph{Conditional Reconstructions}
For the same data point, we show 5 different reconstructions for each model, at different levels of correlation (Fig. \ref{fig:examples_conditional}. In each row, corresponding to a different model, the leftmost image is the input image, the second one is the ground truth. CoVAE is the last model.

\begin{figure}[h]
    \centering
    \begin{subfigure}{0.48\textwidth}
        \centering
        \includegraphics[width=\textwidth]{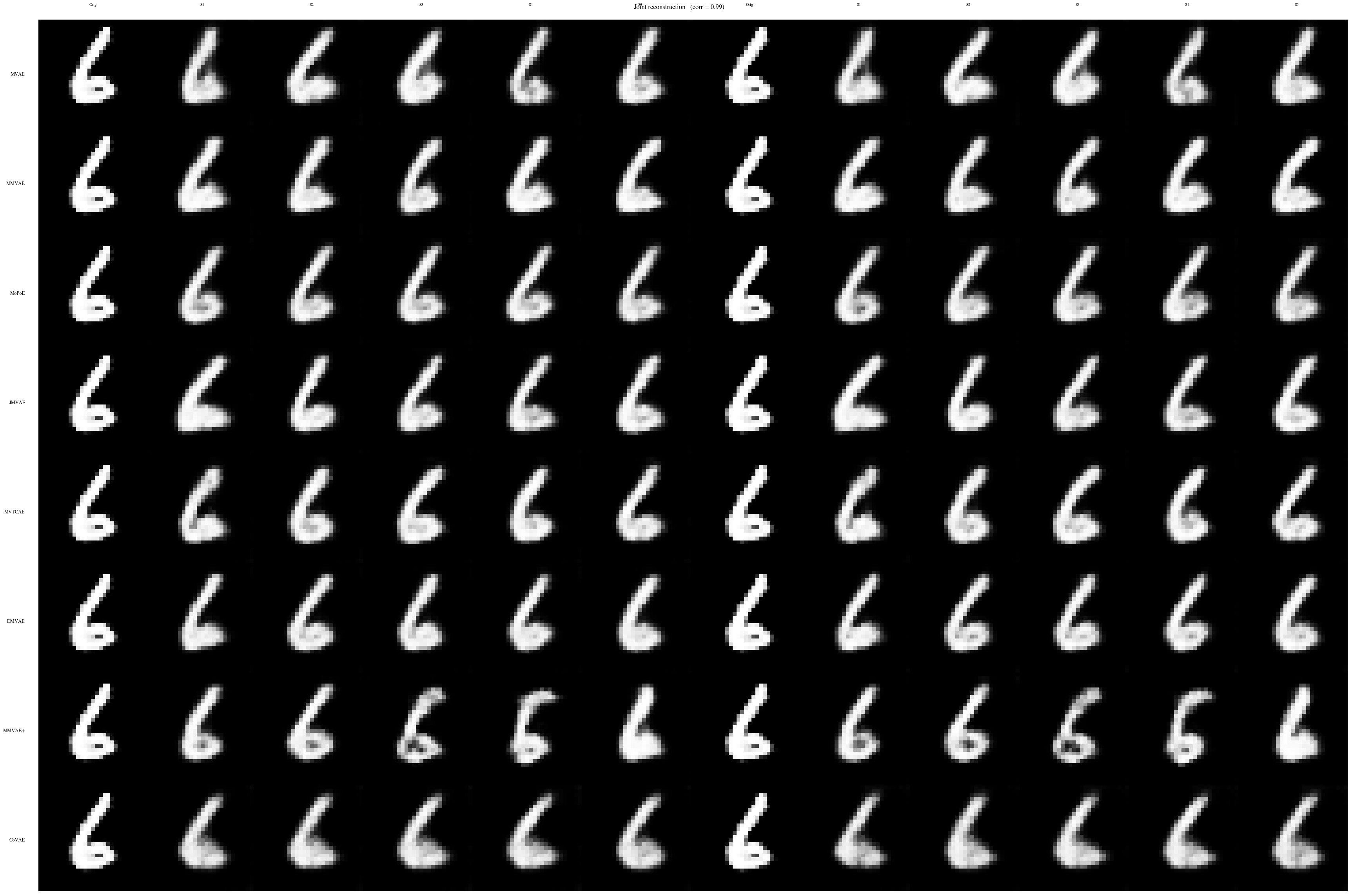}
    \end{subfigure}
    \hfill
    \begin{subfigure}{0.48\textwidth}
        \centering
        \includegraphics[width=\textwidth]{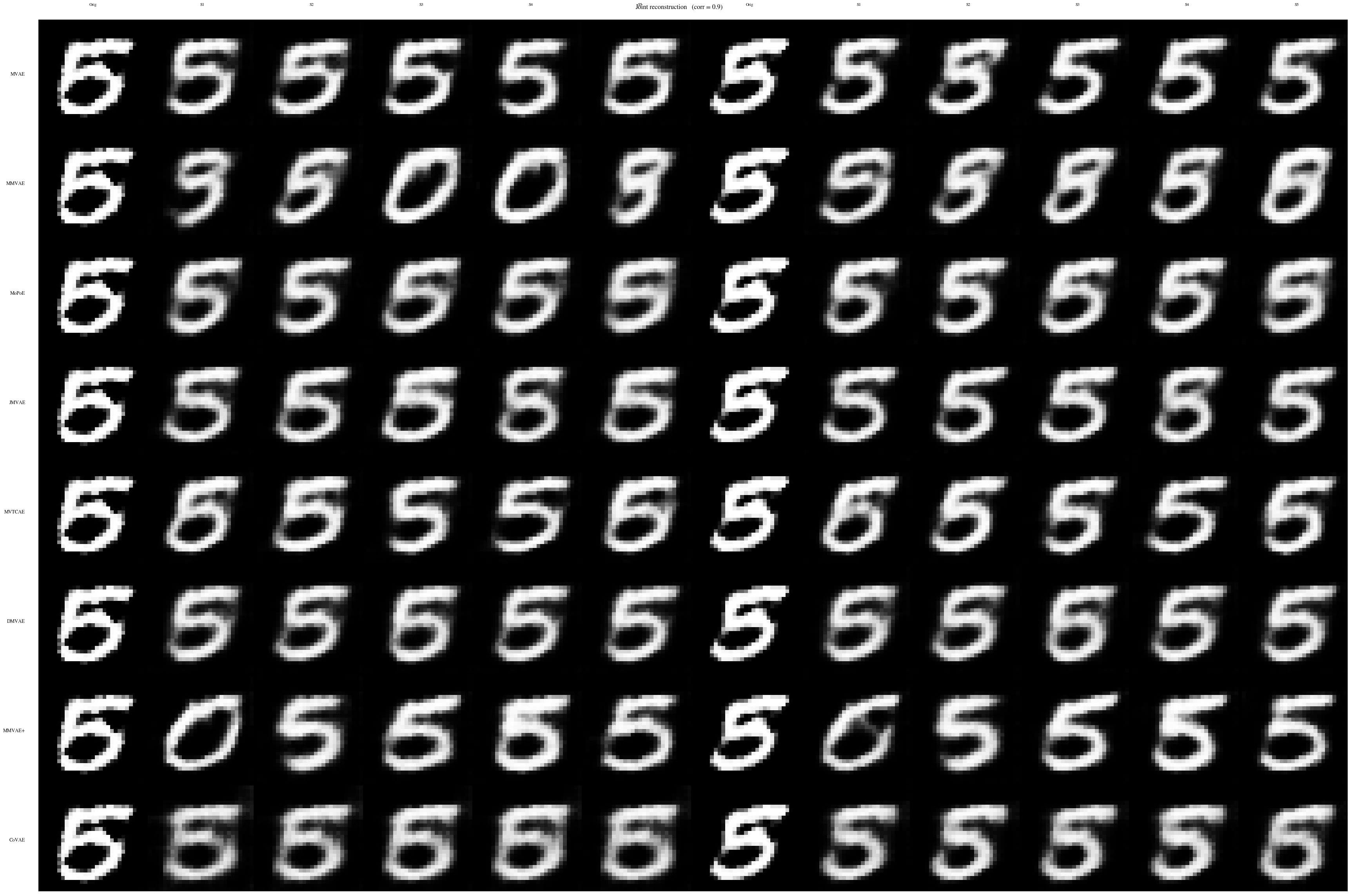}
    \end{subfigure}

    \vspace{0.5em}

    \begin{subfigure}{0.48\textwidth}
        \centering
        \includegraphics[width=\textwidth]{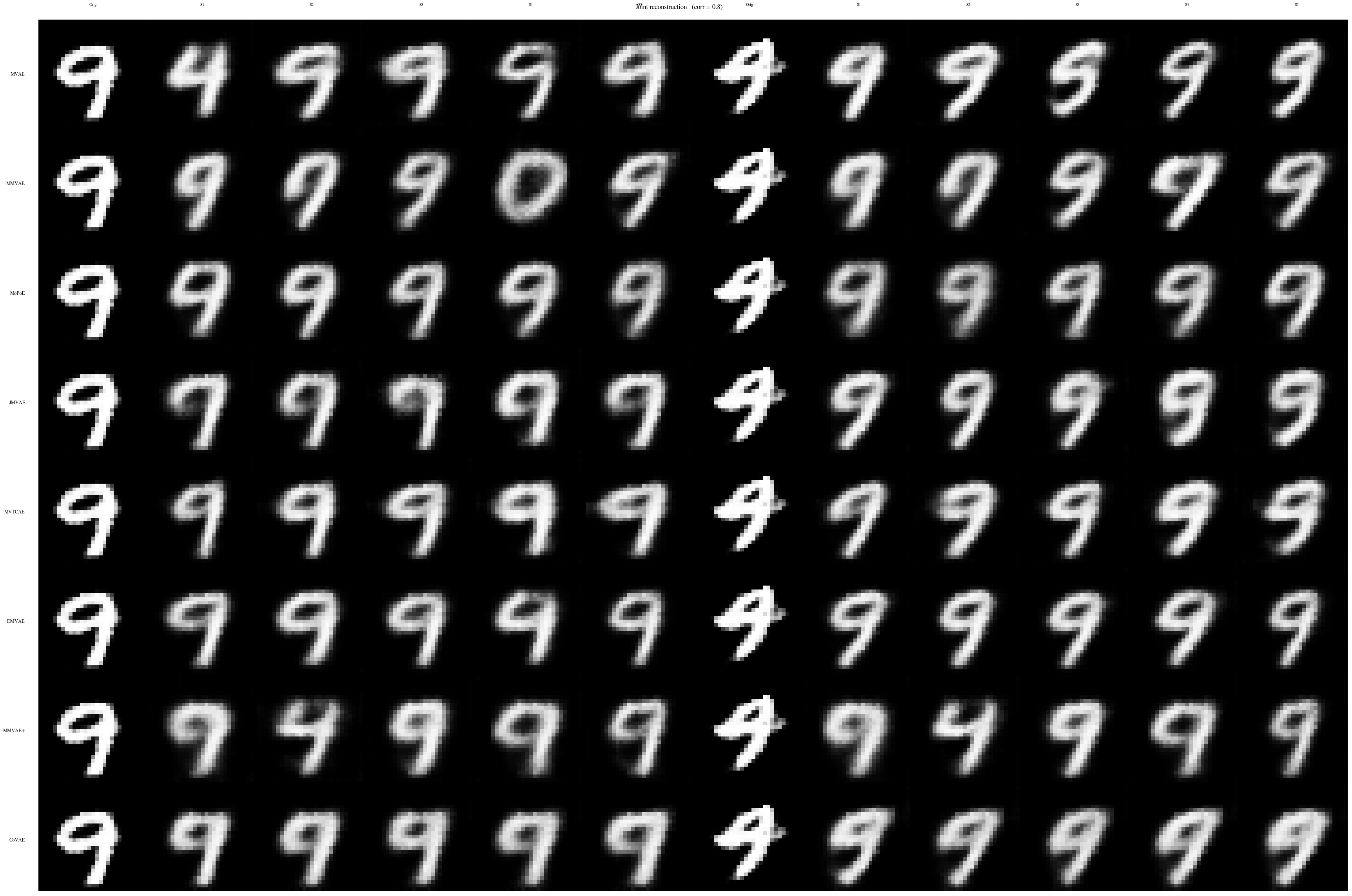}
    \end{subfigure}
    \hfill
    \begin{subfigure}{0.48\textwidth}
        \centering
        \includegraphics[width=\textwidth]{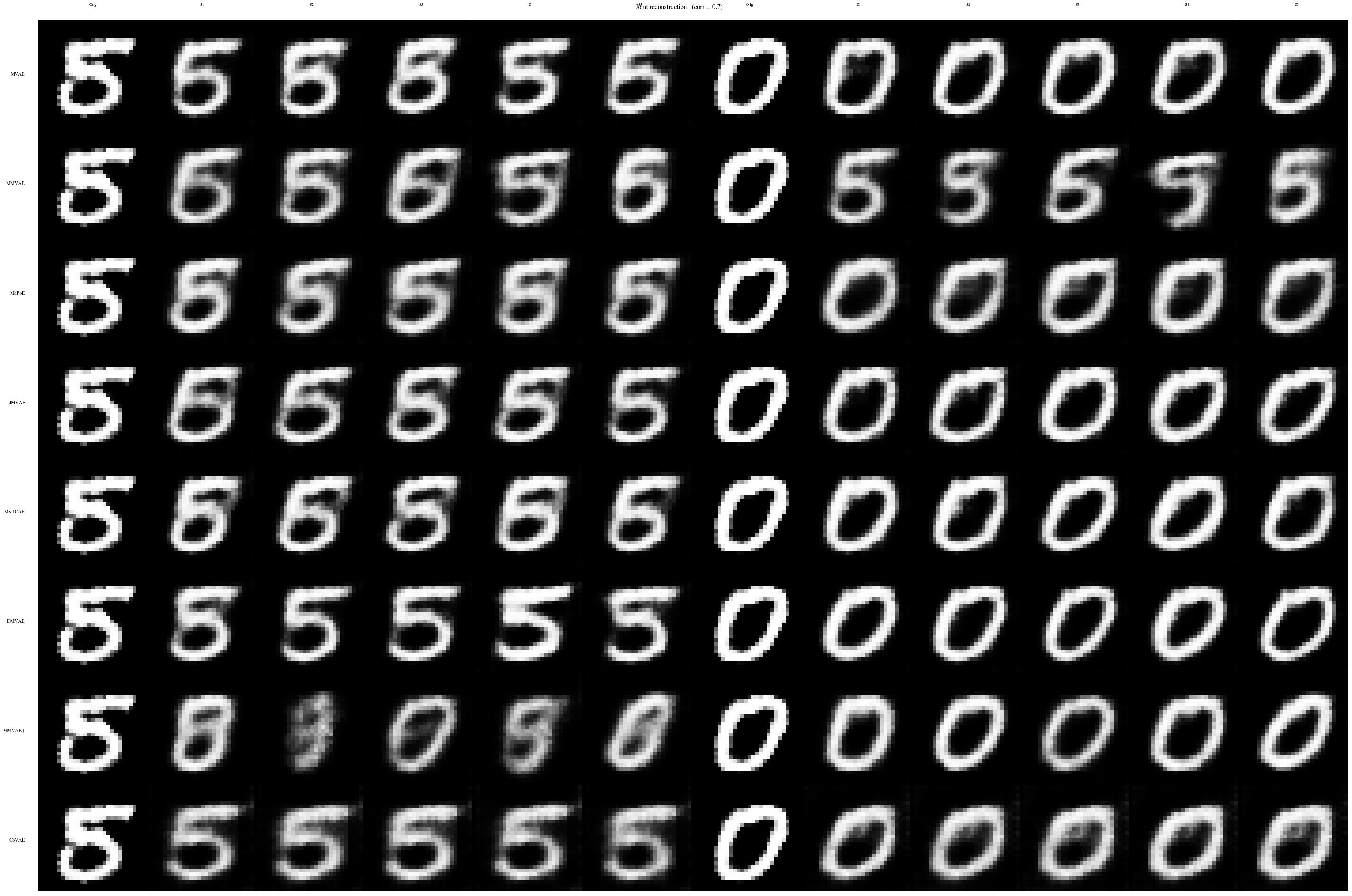}
    \end{subfigure}

    \vspace{0.5em}

    \begin{subfigure}{0.48\textwidth}
        \centering
        \includegraphics[width=\textwidth]{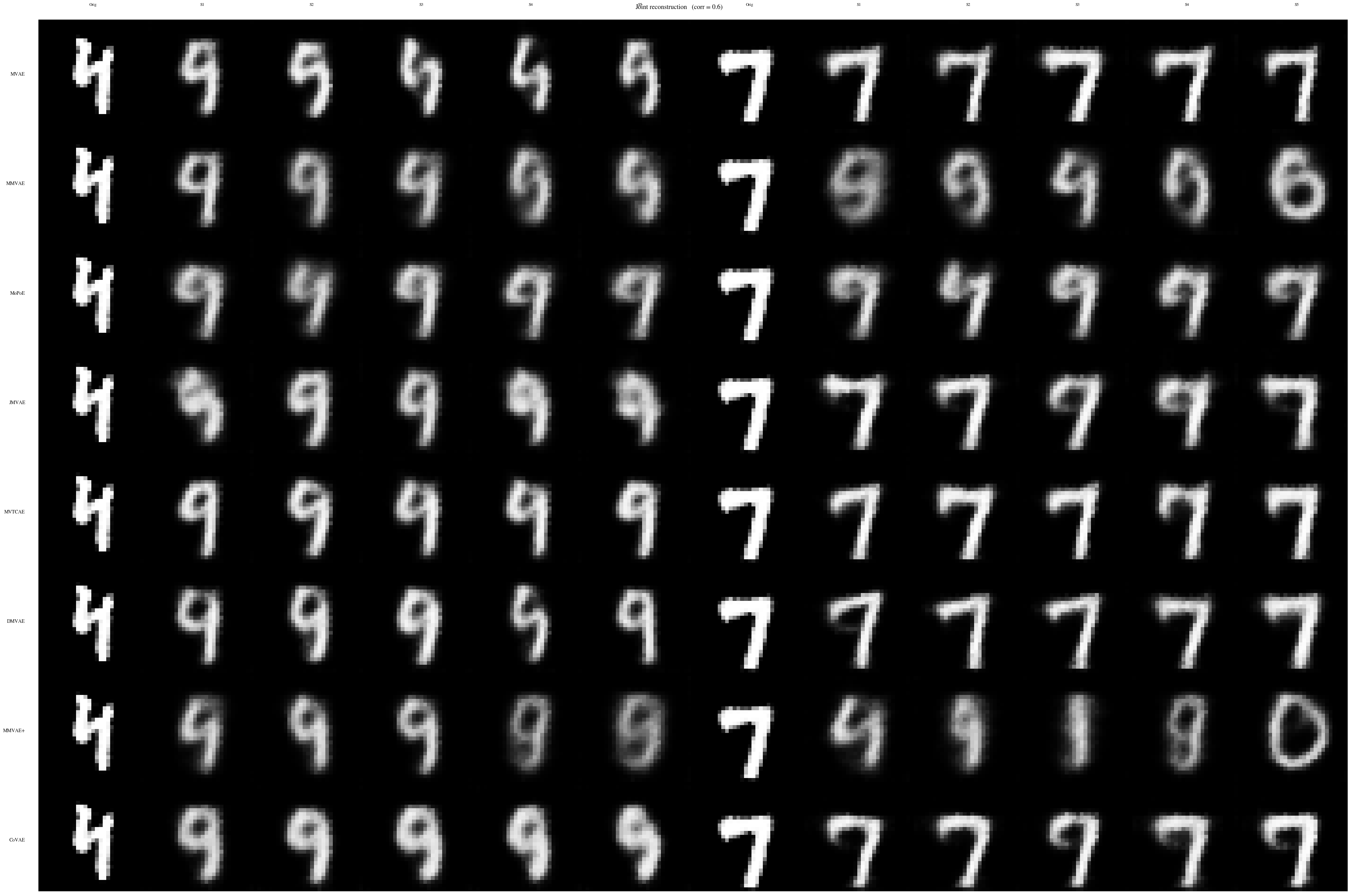}
    \end{subfigure}
    \hfill
    \begin{subfigure}{0.48\textwidth}
        \centering
        \includegraphics[width=\textwidth]{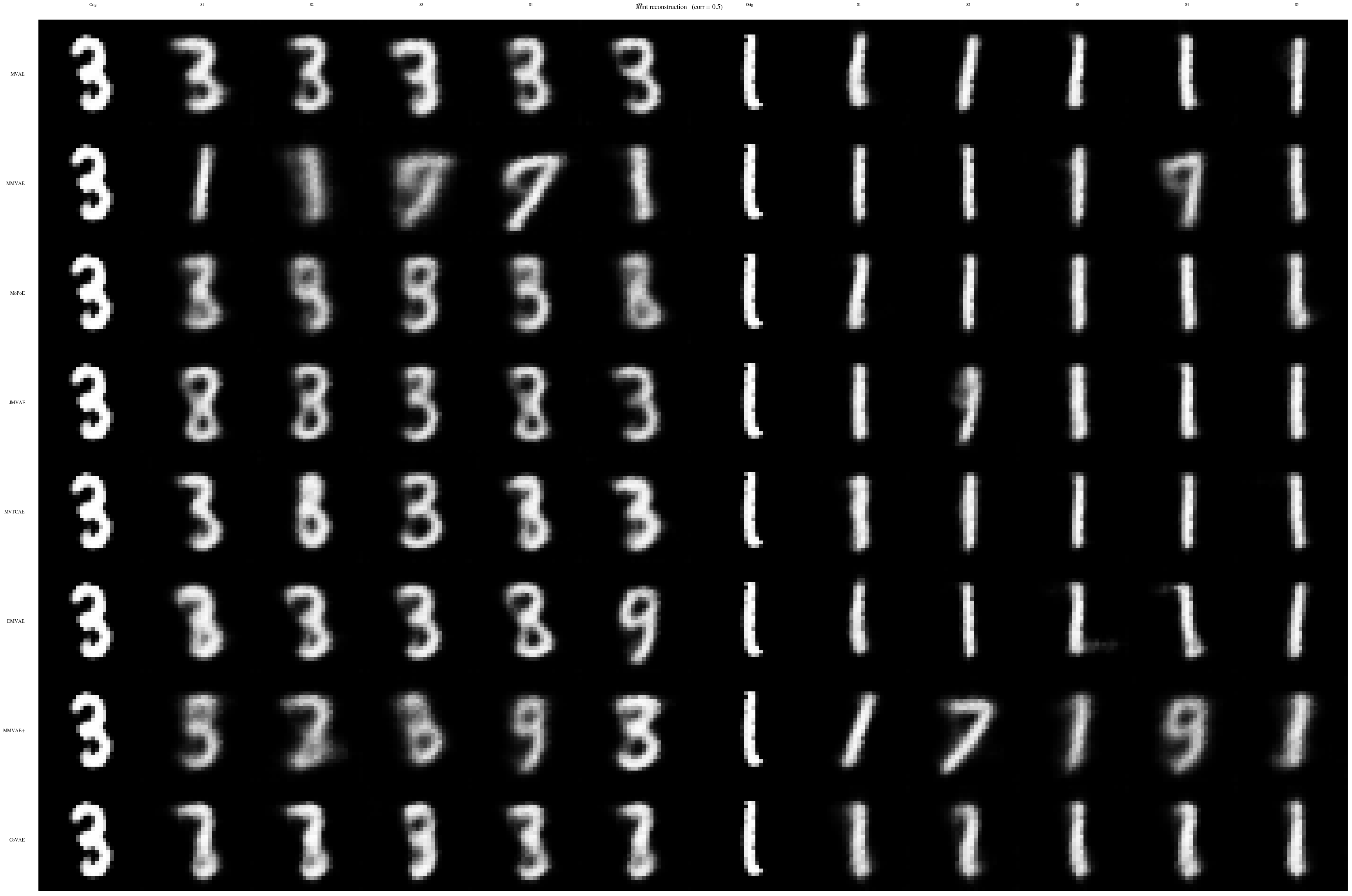}
    \end{subfigure}

    \caption{Joint reconstructions for all models tested at different correlation levels}
    \label{fig:examples_joint}
\end{figure}

\paragraph{Joint Reconstructions}
In this case (Fig.\ref{fig:examples_joint}), the first image of every row represents the input for modality one, and the seventh image the input for modality 2.

\end{document}